\documentclass[journal,transmag]{IEEEtran}

\usepackage{times}
\usepackage{epsfig}
\usepackage{graphicx}
\usepackage{amsmath}
\usepackage{amssymb}

\usepackage{multirow}
\usepackage{booktabs}
\usepackage{bm}
\usepackage{colortbl}

\usepackage{cite}
\usepackage{makecell}
\usepackage{boldline}
\setcellgapes{3pt}
\usepackage{arydshln}
\usepackage{tabu}
\usepackage{subcaption}
\usepackage[pagebackref=true,breaklinks=true,colorlinks,bookmarks=false]{hyperref}

\begin{document}

%%%%%%%%% TITLE
\title{STAR: A Structure and Texture Aware Retinex Model
}

\author{
\IEEEauthorblockN{
Jun Xu\IEEEauthorrefmark{1},
Yingkun Hou\IEEEauthorrefmark{3},
Dongwei Ren\IEEEauthorrefmark{2},
Li Liu\IEEEauthorrefmark{4}, 
Fan Zhu\IEEEauthorrefmark{4}, 
Mengyang Yu\IEEEauthorrefmark{4},
Haoqian Wang\IEEEauthorrefmark{5,6}, and
Ling Shao\IEEEauthorrefmark{4}
%,~\IEEEmembership{Senior Member,~IEEE}
}
\IEEEauthorblockA{\IEEEauthorrefmark{1}College of Computer Science, Nankai University, Tianjin, China}
\IEEEauthorblockA{\IEEEauthorrefmark{2}
College of Intelligence and Computing, Tianjin University, Tianjin, China}
\IEEEauthorblockA{\IEEEauthorrefmark{3}
School of Information Science and Technology, Taishan University, Taian, China}
\IEEEauthorblockA{\IEEEauthorrefmark{4}
Inception Institute of Artificial Intelligence, Abu Dhabi, United Arab Emirates}
\IEEEauthorblockA{\IEEEauthorrefmark{5}Tsinghua Shenzhen International Graduate School, Tsinghua University, Shenzhen, China}internatio
\IEEEauthorblockA{\IEEEauthorrefmark{6}Shenzhen Institute of Future Media Technology, Shenzhen, China}
\thanks{Corresponding Author:\ Haoqian Wang (Email:\ wanghaoqian@tsinghua.edu.cn).}
}

\author{
Jun Xu,
Yingkun Hou,
Dongwei Ren,
Li Liu, 
Fan Zhu, 
Mengyang Yu,
Haoqian Wang,
Ling Shao
\thanks{
This work is partially funded by the Major Project for New Generation of AI under Grant 2018AAA01004, in part by the National Natural Science Foundation of China (No. 61831014, 61929104) and the Shenzhen Science and Technology Project under Grant (JCYJ20170817161916238, JCYJ20180508152042002, GGFW2017040714161462).\ 
The Corresponding author is Prof. Haoqian Wang (Email: wanghaoqian@tsinghua.edu.cn).\
Jun Xu is with TKLNDST, College of Computer Science, Nankai University, Tianjin, China.\
Yingkun Hou is with School of Information Science and Technology, Taishan University, Taian, China.\
Dongwei Ren is with College of Intelligence and Computing, Tianjin University, Tianjin, China.\
Li Liu, Fan Zhu, and Mengyang Yu are with Inception Institute of Artificial Intelligence (IIAI), Abu Dhabi, UAE.\
Haoqian Wang is with the Tsinghua Shenzhen International Graduate School, and also with Shenzhen Institute of Future Media Technology, Shenzhen 518055, China.\
Ling Shao is with the Inception Institute of Artificial Intelligence, Abu Dhabi, UAE, and also with the Mohamed bin Zayed University of Artificial Intelligence, Abu Dhabi, UAE.\ %
}
}

% Figs 5,6,7,8 should be modified by higher resolution images

\maketitle

%%%%%%%%% ABSTRACT
\begin{abstract}
Retinex theory is developed mainly to decompose an image into the illumination and reflectance components by analyzing local image derivatives.\ In this theory, larger derivatives are attributed to the changes in reflectance, while smaller derivatives are emerged in the smooth illumination.\ In this paper, we utilize exponentiated local derivatives (with an exponent $\gamma$) of an observed image to generate its structure map and texture map.\ The structure map is produced by been amplified with $\gamma>1$, while the texture map is generated by been shrank with $\gamma<1$.\ To this end, we design exponential filters for the local derivatives, and present their capability on extracting accurate structure and texture maps, influenced by the choices of exponents $\gamma$.\ The extracted structure and texture maps are employed to regularize the illumination and reflectance components in Retinex decomposition.\ A novel Structure and Texture Aware Retinex (STAR) model is further proposed for illumination and reflectance decomposition of a single image.\ We solve the STAR  model by an alternating optimization algorithm.\ Each sub-problem is transformed into a vectorized least squares regression, with closed-form solutions.\ Comprehensive experiments on commonly tested datasets demonstrate that, the proposed STAR model produce better quantitative and qualitative performance than previous competing methods, on illumination and reflectance decomposition, low-light image enhancement, and color correction.\ The code is publicly available at \url{https://github.com/csjunxu/STAR}.
\end{abstract}

% BODY TEXT
\section{Introduction}
\label{sec:intro}

\begin{figure}
\centering
\begin{minipage}[t]{0.15\textwidth}
\begin{subfigure}[t]{1\linewidth}
\raisebox{-0.15cm}{\includegraphics[width=1\textwidth]{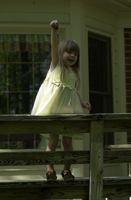}}
\centering{\scriptsize (a) Input }
\end{subfigure}
\begin{subfigure}[b]{1\linewidth}
\raisebox{-0.15cm}{\includegraphics[width=1\textwidth]{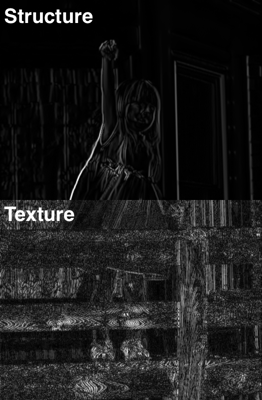}}
\centering{\scriptsize (d) Structure/Texture }
\end{subfigure}
\end{minipage}
\begin{minipage}[t]{0.15\textwidth}
\begin{subfigure}[t]{1\linewidth}
\raisebox{-0.15cm}{\includegraphics[width=1\textwidth]{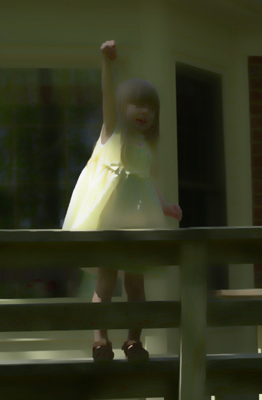}}
\centering{\scriptsize (b) Illumination}
\end{subfigure}
\begin{subfigure}[b]{1\linewidth}
\raisebox{-0.15cm}{\includegraphics[width=1\textwidth]{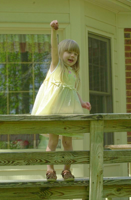}}
\centering{\scriptsize (e) Enhancement}
\end{subfigure}
\end{minipage}
\begin{minipage}[t]{0.15\textwidth}
\begin{subfigure}[t]{1\linewidth}
\raisebox{-0.15cm}{\includegraphics[width=1\textwidth]{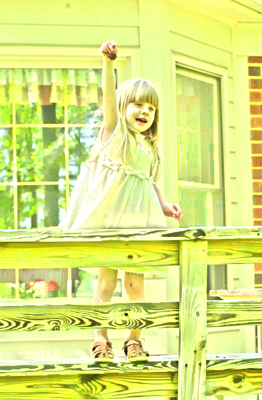}}
\centering{\scriptsize (c) Reflectance}
\end{subfigure}
\begin{subfigure}[b]{1\linewidth}
\raisebox{-0.15cm}{\includegraphics[width=1\textwidth]{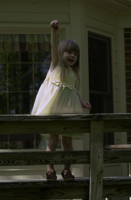}}
\centering{\scriptsize (f) Color Correction}
\end{subfigure}
\end{minipage}
\caption{
An example to illustrate the applications of the proposed STAR model based on Retinex theory.\ (a) The input low-light and color-distorted image;\ (b) the estimated illumination component of (a);\ (c) the estimated reflectance component of (a);\ (d) the extracted structure and texture maps (half each) of (a);\ (e) the illumination enhanced low-light image of (a);\ (f) the color corrected image of (a).}
\label{f-example}
\end{figure}

\IEEEPARstart{T}{he}
Retinex theory developed by Land and McCann~\cite{land1971lightness,land1977retinex} models the color perception of human vision on natural scenes.\ It can be viewed as a fundamental theory for intrinsic image decomposition problem~\cite{barrow1978recovering}, which aims at decomposing an image into illumination and reflectance (or shading) components.\ A simplified Retinex model involves decomposing an observed image $\bm{O}$ into an illumination component $\bm{I}$ and a reflectance component $\bm{R}$ via $\bm{O}=\bm{I}\odot\bm{R}$, where $\odot$ denotes the element-wise multiplication.\ The illumination $\bm{I}$ expresses the color of the light striking the surfaces of objects in the scene $\bm{O}$, while the reflectance $\bm{R}$ reflects the painted color of the surfaces of objects in $\bm{O}$~\cite{closed2012}.\ Retinex theory has been applied in many image processing tasks, such as low-light image enhancement~\cite{closed2012,npe2013,lime2017} and color correction~\cite{wvm2016,jiep2017} (please refer to Figure~\ref{f-example} for an example).

The Retinex theory introduces a useful \textsl{property of derivatives}~\cite{land1971lightness,land1977retinex,closed2012}:\ larger derivatives are often attributed to the changes in reflectance, while smaller derivatives are likely from the smooth illumination.\ With this property, the Retinex decomposition can be performed by classifying the image gradients into the reflectance component and the illumination one~\cite{kimmel2003}.\ However, binary classification of image gradients is unreliable since reflectance and illumination changes will coincide in an intermediate region~\cite{closed2012}.\ Later, several methods are proposed to classify the edges or edge junctions, instead of gradients, according to some trained classifiers~\cite{bell2001learning,tappen2005}.\ However, it is quite challenging to train classifiers considering all possible ranges of illumination and reflectance configurations.\ Besides, though these methods explicitly utilize the \textsl{property of derivatives}, they perform Retinex decomposition by analyzing the gradients of a scene~\cite{Cheng2019} in a \textsl{local} manner, while ignoring the \textsl{global} consistency of the structure in that scene.\ To alleviate this problem, several methods~\cite{closed2012,npe2013,wvm2016} perform global decomposition with the consideration of different regularization.\ However, these methods ignore the \textsl{property of derivatives} and cannot separate well illumination and reflectance components.\ 

In this paper, we propose to utilize exponentiated local derivatives to better exploit the \textsl{property of derivatives} in a global manner.\ The exponentiated derivatives are determined by an introduced exponents $\gamma$ on local derivatives, and generalize the local derivatives to extract global structure and texture maps. Given an observed scene (e.g., Figure~\ref{f-example} (a)), its derivatives are exponentiated by $\gamma$ to generate a structure map (Figure~\ref{f-example} (d) up) when being amplified with $\gamma>1$ and a texture map (Figure~\ref{f-example} (d) down) when being shrank with $\gamma<1$.\ The extracted structure and texture maps are employed to regularize the illumination (Figure~\ref{f-example} (b)) and reflectance (Figure~\ref{f-example} (c)) components in Retinex decomposition, respectively.\ With meaningful structure and texture maps, we propose a Structure and Texture Aware Retinex (STAR) model to accurately estimate the illumination and reflectance components.\ We solve our STAR model by an alternating optimization algorithm~\cite{mcwnnm,pid2020}.\ Each sub-problem is transformed into a vectorized least squares regression with closed-form solutions~\cite{srsc2020}.\ Comprehensive experiments on commonly tested datasets demonstrate that, the proposed STAR model obtains better performance than previous competing methods, on illumination and reflectance decomposition, low-light image enhancement, and color correction.\ 

In summary, the contributions of this work are three-fold:
\begin{itemize}
    \item We propose to utilize exponentially local derivatives to better extract meaningful structure and texture maps.\
    \item We develop a novel Structure and Texture Aware Retinex (STAR) model to accurately estimate the illumination and reflectance components, and exploit the \textsl{property of derivatives} in a global manner.
    \item Experimental on commonly tested datasets demonstrate that our STAR obtains better performance than previous competing methods on Retinex decomposition, low-light image enhancement, and color correction.
\end{itemize}

The remaining paper is organized as follows.\ In \S\ref{sec:related}, we review the related work.\ In \S\ref{sec:stweighting}, we introduce a structure and texture aware weighting scheme for Retinex image decomposition.\ Then we develop a structure and texture aware Retinex model in \S\ref{sec:star}.\ \S\ref{sec:experiments} describes the detailed experiments on Retinex decomposition of illumination and reflectance components.\ \S\ref{sec:appli} presents our STAR model on two other image processing applications: low-light image enhancement and color correction.\ Finally, we conclude this paper in \S\ref{sec:conclusion}.

\begin{figure*}
\centering
\begin{minipage}[t]{0.19\textwidth}
\begin{subfigure}[t]{1\linewidth}
\raisebox{-0.15cm}{\includegraphics[width=1\textwidth]{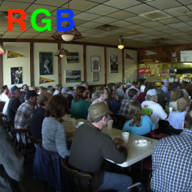}}
\end{subfigure}
\begin{subfigure}[t]{1\linewidth}
\raisebox{-0.15cm}{\includegraphics[width=1\textwidth]{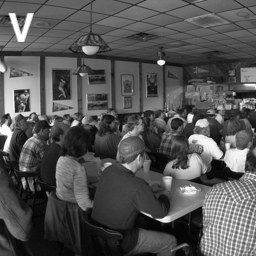}}
\centering{\scriptsize (a) Input }
\end{subfigure}
\end{minipage}
\begin{minipage}[t]{0.19\textwidth}
\begin{subfigure}[b]{1\linewidth}
\raisebox{-0.15cm}{\includegraphics[width=1\textwidth]{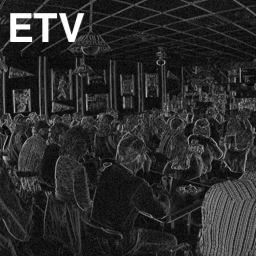}}
\end{subfigure}
\begin{subfigure}[b]{1\linewidth}
\raisebox{-0.15cm}{\includegraphics[width=1\textwidth]{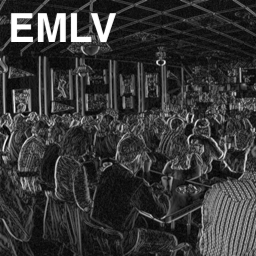}}
\centering{\scriptsize (b) $\gamma=0.5$}
\end{subfigure}
\end{minipage}
\begin{minipage}[t]{0.19\textwidth}
\begin{subfigure}[t]{1\linewidth}
\raisebox{-0.15cm}{\includegraphics[width=1\textwidth]{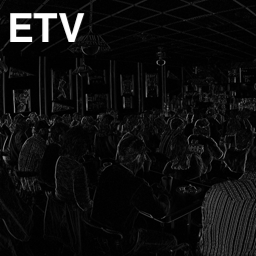}}
\end{subfigure}
\begin{subfigure}[b]{1\linewidth}
\raisebox{-0.15cm}{\includegraphics[width=1\textwidth]{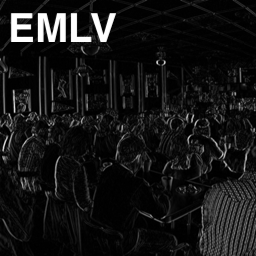}}
\centering{\scriptsize (c) $\gamma=1.0$}
\end{subfigure}
\end{minipage}
\begin{minipage}[t]{0.19\textwidth}
\begin{subfigure}[t]{1\linewidth}
\raisebox{-0.15cm}{\includegraphics[width=1\textwidth]{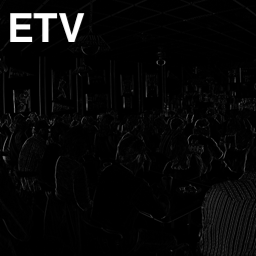}}
\end{subfigure}
\begin{subfigure}[b]{1\linewidth}
\raisebox{-0.15cm}{\includegraphics[width=1\textwidth]{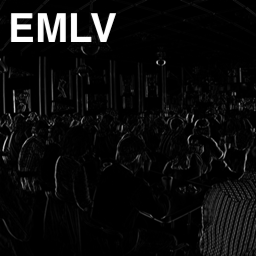}}
\centering{\scriptsize (d) $\gamma=1.5$}
\end{subfigure}
\end{minipage}
\begin{minipage}[t]{0.19\textwidth}
\begin{subfigure}[b]{1\linewidth}
\raisebox{-0.15cm}{\includegraphics[width=1\textwidth]{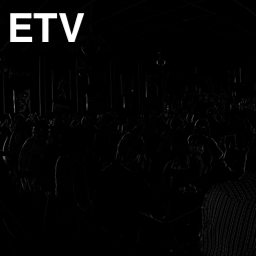}}
\end{subfigure}
\begin{subfigure}[b]{1\linewidth}
\raisebox{-0.15cm}{\includegraphics[width=1\textwidth]{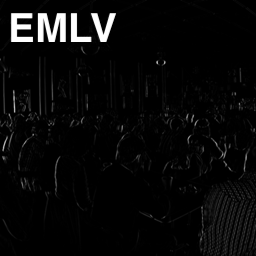}}
\centering{\scriptsize (e) $\gamma=2$}
\end{subfigure}
\end{minipage}
\caption{\linespread{1}\selectfont{
Comparisons on the exponentiated total variation (ETV) and exponentiated mean local variance (EMLV) filters on structure and texture extraction.\ For an input RGB image (a), V refers} to its Value channel in HSV space.}
\label{f-priors}
\end{figure*}

\section{Related Work}
\label{sec:related}

\subsection{Retinex Model}
The Retinex model has been extensively studied in literature~\cite{Land1983,msrcr1997,jiep2017},
%Land1983,brainard1986analysis,frankle1983method,funt2004,HORN1974,morel2010,msrcr1997,kimmel2003,bell2001learning,tappen2005,Provenzi2005,bertalmio2009,palma2009,ng2011total,ma2011,ma2012,li2012,closed2012,npe2013,lime2017,wang2014,liang2015,Morel2015,zosso2015non,wvm2016,jiep2017,li2018structure
which can be roughly divided into classical ones~\cite{brainard1986analysis,funt2004,morel2010} and variational ones~\cite{npe2013,lime2017,jiep2017}.\ Besides, the Retinex decomposition methods can be applied into low-light image enhancement~\cite{bimef,rdgan,Liang_2018_CVPR} and color correction~\cite{lsrs,hu2017fc,ccgans}.

\textbf{Classical Retinex methods} include path based methods~\cite{Land1983,brainard1986analysis,frankle1983method,funt2004}, Partial Differential Equation (PDE) based methods~\cite{morel2010}, and center/surround methods~\cite{msrcr1997}.\ Early path based methods~\cite{Land1983,brainard1986analysis} are developed based on the assumption that, the reflectance component can be computed by the product of ratios along some random paths.\ These methods demand careful parameter tuning and incur high computational costs.\ To improve the efficiency, later path-based methods of~\cite{frankle1983method,funt2004} employ recursive matrix computation techniques to replace previous random path computation.\ However, their performance is largely influenced by the number of recursive iterations, and unstable for real applications.\ PDE based methods~\cite{morel2010} utilize the property that the Retinex solutions satisfy a discrete Poisson equation, which yields an efficient implementation of reflectance estimation using only two Fast Fourier Transformations (FFTs).\ However, the structure of illumination component will be degraded, since gradients derived by a divergence-free vector field often loss piece-wise smoothness.\ The center/surround methods include the famous single-scale Retinex (SSR)~\cite{ssr1997} and multi-scale Retinex with color restoration (MSRCR)~\cite{msrcr1997}.\ These methods simply assume the illumination component to be smooth, and the reflectance component to be non-smooth.\ However, due to lack of a reasonable structure-preserving restriction, MSRCR tends to produce halo artifacts around edges.

\textbf{Variational methods}~\cite{liang2015,wvm2016,li2018structure} have been proposed for Retinex based illumination and reflectance decomposition.\ In~\cite{kimmel2003}, the smooth assumption is introduced into a variational model to estimate the illumination component.\ But this method is slow and ignores to regularize the reflectance.\ Later, an $\ell_{1}$ variational model is proposed in~\cite{ma2011} to focus on estimating the reflectance component.\ But this method ignores to regularize the illumination component.\ The logarithmic transformation is also employed in~\cite{Provenzi2005} as a pre-processing step to suppress the variation of gradient magnitude in bright regions, but the reflectance component estimated with logarithmic regularization tends to be over-smoothed.\ To consider both illumination and reflectance regularizations, a total variation (TV) model based method is proposed in~\cite{ng2011total}.\ But similar to~\cite{Provenzi2005}, the reflectance is over-smoothed due to the side-effect of the logarithmic transformation.\ Recently, Fu \textsl{et al.}~\cite{srie2015} developed a probabilistic method for simultaneous illumination and reflectance estimation (SIRE) in the linear space instead of logarithmic one.\ This method preserves well the details and avoid to over-smooth the reflectance component, when compared to previous methods performed in the logarithmic space.\ To alleviate the detail loss problem of the reflectance component in the logarithmic space, Fu \textsl{et al.}~\cite{wvm2016} proposed a weighted variational model (WVM) to enhance the variation of gradient magnitude in bright regions.\ However, the illumination component may instead be damaged by the unconstrained isotropic smoothness assumption.\ By considering the properties of 3D objects, Cai \textsl{et al.}~\cite{jiep2017} proposed a Joint intrinsic-extrinsic Prior (JieP) model for Retinex decomposition.\ However, this model is prone to over-smoothing both the illumination and reflectance of a scene.\ In~\cite{li2018structure}, Li \textsl{et al.} proposed a Robust Retinex Method (RRM) by considering an additional noise map~\cite{xuaccv2016,xu2019noisy}.\ But this method is effective especially for low-light images accompanied by intensive noise.

\subsection{Intrinsic Image Decomposition}
The Retinex model is in similar spirit with the intrinsic image decomposition model~\cite{Li2014CVPR,ccc2015,ffcc2017}, which decomposes an observed image into Lambertian shading and reflectance (ignoring the specularity).\ The major goal of intrinsic image decomposition is to recover the shading and relectance terms from an observed scene, while the specularity term can be ignored without performance degradation~\cite{Grosse2009}.\ However, the reflectance recovered in this problem usually loses the visual content of the scenes~\cite{lime2017}, and hence can hardly be used for simultaneous illumination and reflectance estimation.\ Therefore, intrinsic image decomposition does not satisfy the purpose of Retinex decomposition for low-light image enhancement, in which the objective is to preserve the visual contents of dark regions as well as keep its visual realism~\cite{lime2017}.\ For more difference between Retinex decomposition and intrinsic image decomposition, please refer to~\cite{lime2017}.

%------------------------------------
\section{Structure and Texture Awareness}
\label{sec:stweighting}

In this section, we first present the simplified Retinex model, and then introduce structure and texture awareness for illumination and reflectance regularization.

%-----------------------------------
\subsection{Simplified Retinex Model}

The Retinex model~\cite{land1977retinex} is a color perception simulation of the human vision system.\ Its physical goal is to decompose an observed image $\bm{O}\in\mathbb{R}^{n\times m}$ into its illumination and reflectance components, i.e.,
\begin{equation}
\label{e-retinex}
\bm{O}
=
\bm{I}\odot\bm{R}
,
\end{equation}
where $\bm{I}\in\mathbb{R}^{n\times m}$ means the illumination component of the scene representing the brightness of objects, $\bm{R}\in\mathbb{R}^{n\times m}$ denotes the surface reflection component of the scene representing its physical characteristics, and $\odot$ means element-wise
multiplication.\ The illumination component $\bm{I}$ and reflectance one $\bm{R}$ can be recovered by alternatively estimating them via
\begin{equation}
\label{e-decomp}
\bm{I}=\bm{O}\oslash\bm{R}
,
\ 
\bm{R}=\bm{O}\oslash\bm{I}
,
\end{equation}
where $\oslash$ means element-wise division.\ In fact, we employ $\bm{I}=\bm{O}\oslash(\bm{R}+\varepsilon)$ and $\bm{R}=\bm{O}\oslash(\bm{I}+\varepsilon)$ to avoid zero denominators, where $\varepsilon=10^{-8}$.\ 

To solve this inverse problem~(\ref{e-decomp}), previous Retinex methods usually employ an objective function that estimates illumination and reflectance components by
\begin{equation}
\label{e-objfun}
\min_{\bm{I},\bm{R}}
\|\bm{O}-\bm{I}\odot\bm{R}\|_{F}^{2}
+
\mathcal{R}_{1}(\bm{I})
+
\mathcal{R}_{2}(\bm{R})
,
\end{equation}
where $\mathcal{R}_{1}$ and $\mathcal{R}_{2}$ are two different regularization functions for illumination $\bm{I}$ and reflectance $\bm{R}$, respectively.\ One implementation choice of $\mathcal{R}_{1}$ and $\mathcal{R}_{2}$ is the total variation (TV)~\cite{rudin1992nonlinear}, which is widely used in previous methods~\cite{ng2011total,wvm2016}.

%-----------------------------------
\subsection{Structure and Texture Estimator}
The Retinex model (\ref{e-retinex}) decomposes an observed scene into its illumination and reflectance components.\ This problem is highly ill-posed, and proper priors of illumination and reflectance should be considered to regularize the solution space.\ Qualitatively speaking, the illumination should be piece-wisely smooth, capturing the structure of the objects in the scene, while the reflectance should present the physical characteristics of the observed scene, capturing its texture information.\ Here, texture refers to the small patterns in object surface, which are similar in local statistics~\cite{WLKT09}.\

Previous structure-texture decomposition methods often enforce the TV regularizers to preserve edges~\cite{ng2011total,liang2015,rtv2012}.\ These TV regularizers simply enforce gradient similarity of the scene and extract the structure of the objects.\ There are two ways for structure-texture decomposition.\ One is to directly derive structure using structure-preserving techniques, such as edge-aware filters~\cite{rgf2014} and optimization based methods~\cite{jiep2017}.\ The other way is to extract structure from the estimated texture weights~\cite{rtv2012}.\ However, these techniques~\cite{rtv2012,rgf2014,jiep2017} are vulnerable to textures and produce ringing effect near edges.\ Moreover, the method~\cite{rtv2012} cannot extract scenes structures with similar appearances to those of the underlying textures.

To better understand the power of these techniques for structure-texture extraction, we study two typical filters.\ The first is the TV filter~\cite{rudin1992nonlinear}, which computes the absolute gradients of an input image as a guidance map:
\begin{equation}
\label{e-tv}
\bm{f}_{TV}(\bm{O})
=
|\nabla\bm{O}|
.
\end{equation}
The second is the mean local variance (MLV)~\cite{jiep2017}, which can also be utilized for structure map estimation:
\begin{equation}
\label{e-mlv}
\bm{f}_{MLV}(\bm{O})
=
\left|\frac{1}{|\Omega|}\sum_{\Omega}\nabla\bm{O}\right|
,   
\end{equation}
where $\Omega$ is the local patch~\cite{Danon2019} around each pixel of $\bm{O}$, $|\Omega|$ denotes the number of elements in $\Omega$, and its size is set as $3\times3$ in all our experiments.\  

\begin{figure}
\centering
\begin{minipage}[t]{0.23\textwidth}
\begin{subfigure}[t]{1\linewidth}
\raisebox{-0.15cm}{\includegraphics[width=1\textwidth]{F-priors/rs8s.png}}
\centering{\scriptsize (a) Input }
\end{subfigure}
\begin{subfigure}[b]{1\linewidth}
\raisebox{-0.15cm}{\includegraphics[width=1\textwidth]{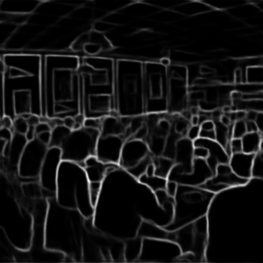}}
\centering{\scriptsize (c) Edge detection of (a) }
\end{subfigure}
\end{minipage}
\begin{minipage}[t]{0.23\textwidth}
\begin{subfigure}[t]{1\linewidth}
\raisebox{-0.15cm}{\includegraphics[width=1\textwidth]{F-priors/rs8ETV1.png}}
\centering{\scriptsize (b) TV filtered (a)}
\end{subfigure}
\begin{subfigure}[b]{1\linewidth}
\raisebox{-0.15cm}{\includegraphics[width=1\textwidth]{F-priors/rs8EMLV1.png}}
\centering{\scriptsize (d) MLV filtered (a)}
\end{subfigure}
\end{minipage}
\caption{Comparison of TV filtered image (b), MLV filtered image (d), and the edge extracted image (c) of the input image (a).\ It can be seen that the TV and MLV filtered images can roughly reflect the structure of the input image.}
\label{f-edges}
\end{figure}

To support that the TV and MLV filters can capture the structure of the scene, we visualize the effect of the two filters performed on extracting the structure/texture from an observed image.\ Here, the input RGB image (Figure~\ref{f-priors} (a), up) is first transformed into the Hue-Saturation-Value (HSV) domain.\ Since the Value (V) channel (Figure~\ref{f-priors} (a), down) reflects the illumination and reflectance information, we process this channel for the input image.\ It can be seen from Figure~\ref{f-priors} (c) that, the TV and MLV filters can basically reflect the main structure of the input image.\ This point can be further validated by comparing the similarity of the two filtered image (Figure~\ref{f-priors} (c)) with the edges extracted for the input image (Figure~\ref{f-priors} (a)).\ To this end, we resort to a recently published edge detection method~\cite{Rcf2019} to extract the main structure of the input image.\ By comparing the TV filtered image (Figure~\ref{f-edges} (b)), MLV filtered image (Figure~\ref{f-edges} (d)), and the edge extracted image (Figure~\ref{f-edges} (c)), we observe that the TV and MLV filtered images already reflect the structure of the input image.\

%-----------------------------------
\subsection{Proposed Structure and Texture Awareness}
Existing TV and MLV filters described in Eqns.~(\ref{e-tv}) and~(\ref{e-mlv}) cannot be directly utilized in our problem, since they are prone to capture structural information.\ As described in Retinex theory~\cite{land1971lightness,land1977retinex}, larger derivatives are attributed to the changes in reflectance, while smaller derivatives are emerged in the smooth illumination.\ Therefore, by exponential growth or decay, these local derivatives will reflect more clearly the corresponding content structure or detailed textures, as has been illustrated in Figure~\ref{f-priors}.\ To this end, we introduce an exponential version of local derivatives for flexible structure and texture estimation.\ Specifically, we add an exponent term to the TV and MLV filtering operations.\ By this way, we can make the two filters more flexible for separate structure and texture extraction.\ Specifically, we propose the exponentiated TV (ETV) filter as
\begin{equation}
\label{e-etvf}
\bm{f}_{ETV}(\bm{O})
=
\bm{f}_{TV}^{\gamma}(\bm{O})
=
|\nabla\bm{O}|^{\gamma}
,
\end{equation}
and the exponentiated MLV (EMLV) filter as
\begin{equation}
\label{e-emlvf}
\bm{f}_{EMLV}(\bm{O})
=
\bm{f}_{MLV}^{\gamma}(\bm{O})
=
\left|\frac{1}{|\Omega|}\sum_{\Omega}\nabla\bm{O}\right|^{\gamma}
,
\end{equation}
where $|\Omega|$ denotes the number of elements in $\Omega$ and $\gamma$ is the exponent determining the sensitivity to the gradients of $\bm{O}$.\ Note that we evaluate the two exponentiated filters Eqns.~(\ref{e-etvf}) and (\ref{e-emlvf}) by visualizing their effects on a test image (i.e., Figure~\ref{f-priors} (a), top).\ This RGB image is first transformed into the Hue-Saturation-Value (HSV) domain, and the decomposition is performed in the Value (V) channel.\ In Figure~\ref{f-priors} (b)-(e), we plot the filtered images for the V channel of the input image.\ It is noteworthy that, with $\gamma=0.5$, the ETV and EMLV filters roughly reveal the textures of the test image, while with $\gamma \in\{1, 1.5, 2\}$, the ETV and EMLV filters tend to extract the structural edges.\

Motivated by this observation, we introduce a structure and texture aware weighting scheme for illumination and reflectance decomposition. Specifically, we set $\bm{I}_{0}=\bm{R}_{0}=\bm{O}^{0.5}$, the ETV based weighting matrix as
\begin{equation}
\label{e-tvf}
\bm{S}_{0}
=
1\oslash(\left|\nabla\bm{I}_{0}\right|^{\gamma_{s}}+\varepsilon)
,\
\bm{T}_{0}
=
1\oslash(\left|\nabla\bm{R}_{0}\right|^{\gamma_{t}}+\varepsilon)
,
\end{equation}
and the EMLV based weighting matrix as:
\begin{equation}
\begin{split}
\label{e-mlvf}
\bm{S}_{0}
&
=
1\oslash(\left|\frac{1}{|\Omega|}\sum_{\Omega}\nabla\bm{I}_{0}\right|^{\gamma_{s}}+\varepsilon)
,
\\
\bm{T}_{0}
&
=
1\oslash(\left|\frac{1}{|\Omega|}\sum_{\Omega}\nabla\bm{R}_{0}\right|^{\gamma_{t}}+\varepsilon)
,
\end{split}
\end{equation}
where $\gamma_{s}>1$ and $\gamma_{t}<1$ are two exponential parameters to adjust the structure and texture awareness for illumination and reflectance decomposition.\ As will be demonstrated in \S\ref{sec:experiments}, the values of $\gamma_{s}$ and $\gamma_{t}$ influence the performance of the Retinex decomposition.\ Due to considering local variance information, the EMLV filter (Eqn.~(\ref{e-mlvf})) can reveal details and preserve structures better than the ETV filter (Figure~\ref{f-priors}).\ This point will also be validated in \S\ref{sec:experiments}.

%------------------------------------------------------------------------
\section{Structure and Texture Aware Retinex Model}
\label{sec:star}

\subsection{Proposed Model}
In this section, we propose a Structure and Texture Aware Retinex (STAR) model to simultaneously estimate the illumination $\bm{I}$ and the reflectance $\bm{R}$ of an observed image $\bm{O}$.\ To make our STAR model as simple as possible, we adopt the TV $\ell_{2}$-norm to regularize the illumination and reflectance components.\ The proposed STAR model is formulated as
\begin{equation}
\label{e-star}
\min_{\bm{I},\bm{R}}
\|\bm{O}-\bm{I}\odot\bm{R}\|_{F}^{2}
+
\alpha
\|\bm{S}_{0}\odot\nabla\bm{I}\|_{F}^{2}
+
\beta
\|\bm{T}_{0}\odot\nabla\bm{R}\|_{F}^{2}
,
\end{equation}
where $\bm{S}_{0}$ and $\bm{T}_{0}$ are the two matrices defined in (\ref{e-mlvf}), indicating the structure map of the illumination and the texture map of the reflectance, respectively.\ The structure should be small enough to preserve the edges of objects in the scene, while large enough to suppress the details (as the inverse of Figure~\ref{f-priors} (d,e)).\ On the other hand, the texture map should be small enough to reveal the details (as the inverse of Figure~\ref{f-priors} (b,c)).\
\begin{table}[t]
\centering
\begin{tabular}{l}
\Xhline{1pt}
\textbf{Algorithm 1}: Solve the STAR Model (\ref{e-star})
\\
\hline
\textbf{Input:} observed image $\bm{O}$, parameters $\alpha,\beta$, 
\\
\quad\quad\quad
maximum iteration number $K$;
\\
\textbf{Initialization:} $\bm{I}_{0}=\bm{O}^{0.5}$, $\bm{R}_{0}=\bm{O}^{0.5}$, set $\bm{S}_{0},\bm{T}_{0}$ by (\ref{e-mlvf});
\\
%\hline
\textbf{for}~($k = 0,...,K-1$)~\textbf{do}
\\
1. 
\quad
Update $\bm{I}_{k+1}$ by Eqn.\ (\ref{e-solvei});
\\
2. 
\quad
Update $\bm{R}_{k+1}$ by Eqn.\ (\ref{e-solver});
\\
\quad\quad
\textbf{if} (Converged)
\\
3.
\quad\quad
\text{Stop};
\\
\quad\quad
\textbf{end if}
\\
\textbf{end for}
\\
%\hline
\textbf{Output:} Estimated illumination $\hat{\bm{I}}$ and reflectance $\hat{\bm{R}}$.
\\
\Xhline{1pt}
\end{tabular}
\end{table}

\subsection{Optimization Algorithm}
Since the objective function (\ref{e-star}) is separable \textsl{w.r.t.} the two variables $\bm{I}$ and $\bm{R}$, it can be solved via an alternative optimization algorithm.\ The two separated sub-problems are convex and alternatively solved.\ We initialize the matrix variables $\bm{I}_{0}=\bm{R}_{0}=\bm{O}^{0.5}$.\ Denote $\bm{I}_{k}$ and $\bm{R}_{k}$ as the illumination and reflectance components at the $k$-th ($k = 0, 1, 2, ..., K$) iteration, respectively, and $K$ is the maximum iteration number.\ By optimizing one variable at a time while fixing the other, we can alternatively update the two variables as follows:
\begin{table}[t]
\centering
\begin{tabular}{l}
\Xhline{1pt}
\textbf{Algorithm 2}: Alternative Updating Scheme
\\
\hline
\textbf{Input:} observed image $\bm{O}$, parameters $\alpha,\beta$, updating number $L$,
\\
\quad\quad\quad
maximum iteration number $K$ defined in Algorithm 1;
\\
\textbf{Initialization:} estimated $\hat{\bm{I}}^{0}=\hat{\bm{I}},\hat{\bm{R}}^{0}=\hat{\bm{R}}$ by \textbf{Algorithm 1};
\\
%\hline
\textbf{for}~($l = 0,...,L-1$)~\textbf{do}
\\
1.
\quad
Update
$\bm{S}_{l+1}
=
(|\frac{1}{|\Omega|}\sum_{\Omega}\nabla\hat{\bm{I}}^{l}|^{\gamma_{s}}+\varepsilon)^{-1}$;
\\
2.
\quad
Update $\bm{T}_{l+1}
=
(|\frac{1}{|\Omega|}\sum_{\Omega}\nabla\hat{\bm{R}}^{l}|^{\gamma_{t}}+\varepsilon)^{-1}$; 
\\
3. 
\quad
Solve the STAR model~(\ref{e-star}) and obtain $\hat{\bm{I}}^{l+1}$ and
\\
\quad\quad
$\hat{\bm{R}}^{l+1}$ by \textbf{Algorithm 1};
\\
\quad\quad
\textbf{if} (Converged)
\\
4.
\quad\quad\quad
\text{Stop};
\\
\quad\quad
\textbf{end if}
\\
\textbf{end for}
\\
%\hline
\textbf{Output:} Final illuminance $\hat{\bm{I}}^{L}$ and reflectance $\hat{\bm{R}}^{L}$.
\\
\Xhline{1pt}
\end{tabular}
\end{table}

\noindent
\textbf{a) Update $\bm{I}$ while fixing $\bm{R}$.}\ With $\bm{R}_{k}$ in the $k$-th iteration, the optimization problem with respect to $\bm{I}$ becomes:
\begin{equation}
\label{e-upI}
\bm{I}_{k+1}
=
\arg\min_{\bm{I}}
\|\bm{O}-\bm{I}\odot\bm{R}_{k}\|_{F}^{2}
+
\alpha 
\|\bm{S}_{0}\odot\nabla\bm{I}\|_{F}^{2}
.
\end{equation}
To solve the problem~(\ref{e-upI}), we reformulate it into a vectorized format.\ To this end, with the vectorization operator $\text{vec}(\cdot)$, we denote vectors $\bm{o}=\text{vec}(\bm{O})$, $\bm{i}=\text{vec}(\bm{I})$,
$\bm{r}_{k}=\text{vec}(\bm{R}_{k})$,
$\bm{s}_{0}=\text{vec}(\bm{S}_{0})$, which are of length $nm$.\ Denote by $\bm{G}$ the Toeplitz matrix from the discrete gradient operator with forward difference, then we have $\bm{G}\bm{i}=\text{vec}(\nabla\bm{I})$.\ Denote by $\bm{D}_{\bm{r}_{k}}=\text{diag}(\bm{r}_{k})$, $\bm{D}_{\bm{s}_{0}}=\text{diag}(\bm{s}_{0})\in\mathbb{R}^{nm\times nm}$ the matrices with $\bm{r}_{k},\bm{s}_{0}$ lying on the main diagonals, respectively.\ Then, the problem~(\ref{e-upI}) is transformed into a standard least squares regression problem:
\begin{equation}
\label{e-upi}
\bm{i}_{k+1}
=
\arg\min_{\bm{i}}
\|\bm{o}-\bm{D}_{\bm{r}_{k}}\bm{i}\|_{2}^{2}
+
\alpha 
\|\bm{D}_{\bm{s}_{0}}\bm{G}\bm{i}\|_{2}^{2}
.
\end{equation}
By differentiating problem~(\ref{e-upi}) with respect to $\bm{i}$, and setting the derivative to $\bm{0}$, we have the following solution
\begin{equation}
\label{e-solvei}
\bm{i}_{k+1}
=
(
\bm{D}_{\bm{r}_{k}}^{\top}\bm{D}_{\bm{r}_{k}}
+
\alpha                            
\bm{G}^{\top}
\bm{D}_{\bm{s}_{0}}^{\top}
\bm{D}_{\bm{s}_{0}}
\bm{G}
)^{-1}
\bm{D}_{\bm{r}_{k}}^{\top}\bm{o}
.
\end{equation}
We then reformulate the obtained $\bm{i}_{k+1}$ into matrix format via the inverse vectorization $\bm{I}_{k+1}=\text{vec}^{-1}(\bm{i}_{k+1})$.

\noindent
\textbf{b) Update $\bm{R}$ while fixing $\bm{I}$.}\ After acquiring $\bm{I}_{k+1}$ from the solution (\ref{e-upI}), the optimization problem~(\ref{e-star}) with respect to $\bm{R}$ is similar to that of $\bm{I}$:
\begin{equation}
\label{e-upR}
\bm{R}_{k+1}
=
\arg\min_{\bm{R}}
\|\bm{O}-\bm{I}_{k+1}\odot\bm{R}\|_{F}^{2}
+
\beta
\|\bm{T}_{0}\odot\nabla\bm{R}\|_{F}^{2}
.
\end{equation}
Similarly, we reformulate the problem~(\ref{e-upR}) into a vectorized format.\ Additionally, we denote $\bm{r}$$=$$\text{vec}(\bm{R})$, $\bm{t}_{0}$$=$$\text{vec}(\bm{T}_{0})$.\  which are of length $nm$.\ We also have $\bm{G}\bm{r}=\text{vec}(\nabla\bm{R})$.\ Denote by $\bm{D}_{\bm{i}_{k+1}}=\text{diag}(\bm{i}_{k+1})$, $\bm{D}_{\bm{t}_{0}}=\text{diag}(\bm{t}_{0})\in\mathbb{R}^{nm\times nm}$ the matrices with $\bm{r}_{k},\bm{s}_{0}$ lying on the main diagonals, respectively.\ Then, the problem~(\ref{e-upR}) is also transformed into a standard least squares problem:
\begin{equation}
\label{e-upr}
\bm{r}_{k+1}
=
\arg\min_{\bm{r}}
\|\bm{o}-\bm{D}_{\bm{i}_{k+1}}\bm{r}\|_{2}^{2}
+
\beta 
\|\bm{D}_{\bm{t}_{0}}\bm{G}\bm{r}\|_{2}^{2}
.
\end{equation}
By differentiating the problem~(\ref{e-upr}) with respect to $\bm{r}$, and setting the derivative to $\bm{0}$, we have the following solution
\begin{equation}
\label{e-solver}
\bm{r}_{k+1}
=
(
\bm{D}_{\bm{i}_{k+1}}^{\top}\bm{D}_{\bm{i}_{k+1}}
+
\beta                            
\bm{G}^{\top}
\bm{D}_{\bm{t}_{0}}^{\top}
\bm{D}_{\bm{t}_{0}}
\bm{G}
)^{-1}
\bm{D}_{\bm{i}_{k+1}}^{\top}\bm{o}
.
\end{equation}
We then reformulate the obtained $\bm{r}_{k+1}$ into the matrix format via inverse vectorization $\bm{R}_{k+1}=\text{vec}^{-1}(\bm{r}_{k+1})$.

The above alternative algorithm are repeated until the convergence condition is satisfied or the number of iterations exceeds a preset threshold.\ The convergence condition of the alternative optimization algorithm is: $\|\bm{I}_{k+1}-\bm{I}_{k}\|_{F}/\|\bm{I}_{k}\|_{F}\le \varepsilon$ or $\|\bm{R}_{k+1}-\bm{R}_{k}\|_{F}/\|\bm{R}_{k}\|_{F}\|\le \varepsilon$ is satisfied, or the maximum iteration number $K$ is achieved.\ We set $\varepsilon=10^{-2}$ and $K=20$ in our experiments.\ Our STAR model (\ref{e-star}) can be efficiently solved since there are only two variables in problem (\ref{e-star}) and each sub-problem has closed-form solution.\

\begin{figure}[h!]
\vspace{-1mm}
\includegraphics[width=0.45\textwidth]{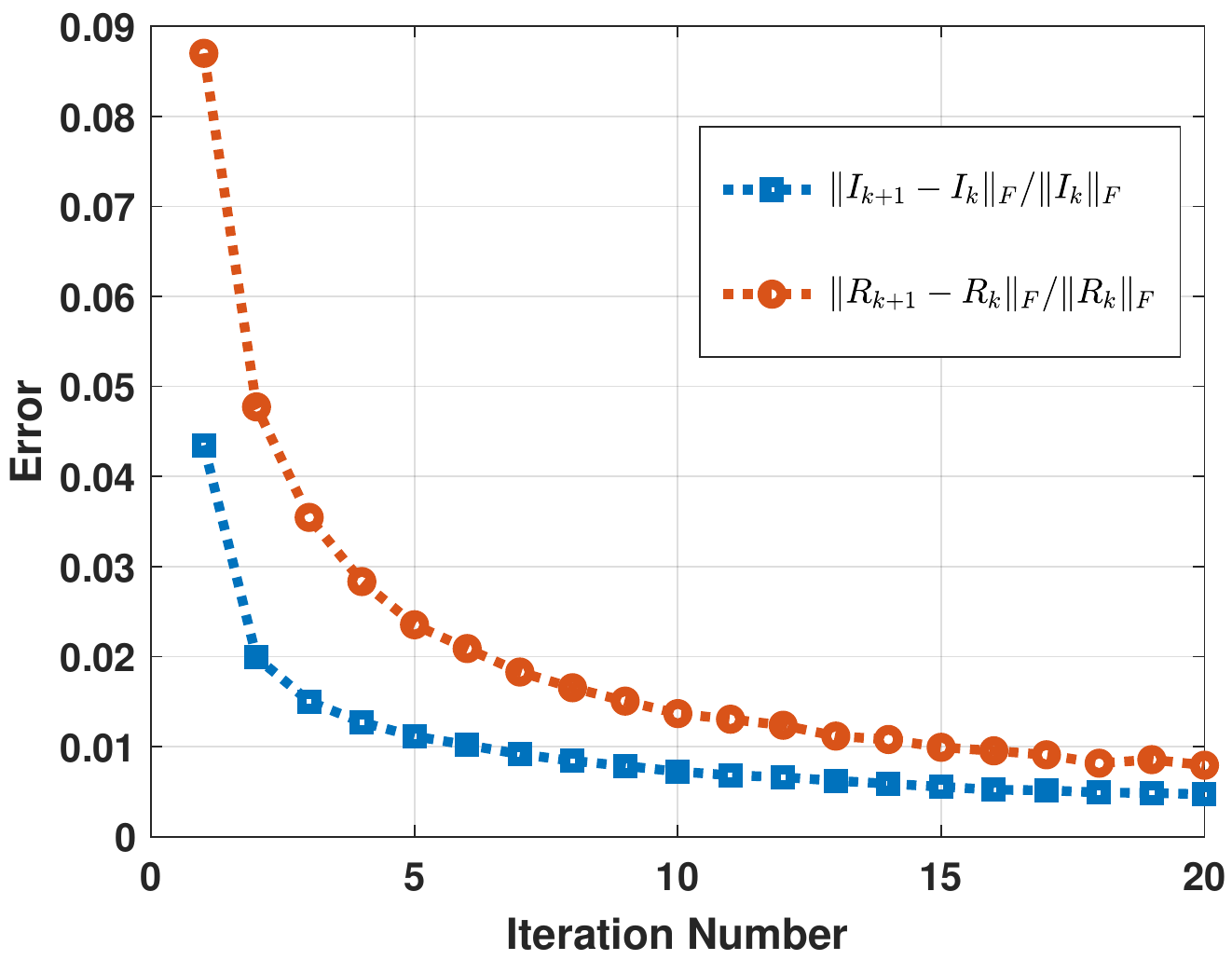}
\vspace{-2mm}
\caption{The average convergence curves of $\|\bm{I}_{k+1}-\bm{I}_{k}\|_{F}/\|\bm{I}_{k}\|_{F}$ (\textcolor[rgb]{0.00,0.45,0.74}{blue} line) and $\|\bm{R}_{k+1}-\bm{R}_{k}\|_{F}/\|\bm{R}_{k}\|_{F}$ (\textcolor[rgb]{0.85,0.33,0.10}{orange} line) of our STAR on the 35 images collected from~\cite{srie2015,wvm2016,lime2017,jiep2017}.}
\label{F-Convergence}
\vspace{-3mm}
\end{figure}

\textbf{Convergence Analysis}.\ The convergence of Algorithm 1 can be guaranteed since the overall objective function (\ref{e-star}) is convex with a global optimal solution.\ In Figure~\ref{F-Convergence}, we plot the \{average convergence curves of the errors of $\|\bm{I}_{k+1}-\bm{I}_{k}\|_{F}/\|\bm{I}_{k}\|_{F}$ or $\|\bm{R}_{k+1}-\bm{R}_{k}\|_{F}/\|\bm{R}_{k}\|_{F}\|$ on the 35 low-light images collected from~\cite{srie2015,wvm2016,lime2017,jiep2017}.\ One can see that either of them is reduced to less than $\varepsilon=0.01$ in 10 iterations.

\subsection{Updating Structure and Texture Awareness}
Until now, we have obtained the decomposition of $\bm{O}=\bm{I}\odot\bm{R}$.\ To achieve better estimation on illumination and reflectance, we update the structure and texture aware maps $\bm{S}$ and $\bm{T}$, and then solve the renewed problem (\ref{e-star}).\ The alternative updating of ($\bm{S}$, $\bm{T}$) and ($\bm{I}$, $\bm{R}$) are repeated for $L$ iterations.\ We set $L=4$ to balance the speed-accuracy trade-off of the proposed STAR model in our experiments.\ We summarize the updating procedures in Algorithm 2.
% The convergence condition of for this algorithm is: $\|\bm{S}_{l+1}-\bm{S}_{l}\|\le\varepsilon$ and $\|\bm{T}_{l+1}-\bm{T}_{l}\|\le\varepsilon$ are simultaneously satisfied, or the maximum updating iteration number $L$ is achieved.\

\noindent
\textbf{Complexity Analysis}.\
Now we discuss the complexity analysis of the proposed Algorithms 1 and 2.\ Assume that the input image is of size $n\times m$.\ In Algorithm 1, the costs for updating $\bm{I}$ and $\bm{R}$ are both $\mathcal{O}(nmK)$ due to the diagonalization operations, where $K$ is the number of iterations in Algorithm 1.\ In Algorithm 2, the costs for updating $\bm{S}$ and $\bm{T}$ are also  $\mathcal{O}(nmKL)$, where $L$ is the number of updating in Algorithm 2.\ As such, the overall complexity of our STAR for Retinex decomposition is $\mathcal{O}(nmKL)$.

%------------------------------------------------------------------------
\section{Experiments}
\label{sec:experiments} 
In this section, we evaluate the qualitative and quantitative performance of the proposed Structure and Texture Aware Retinex (STAR) model on Retinex decomposition (\S\ref{sec:decom}).\ In \S\ref{sec:validation}, we also perform an ablation study on illumination and reflectance decomposition to gain deeper insights into the proposed STAR Retinex model.\ All these experiments are run on a Huawei Matebook X Pro laptop with an Intel Core i5 8265U CPU and 8GB memory.

\subsection{Implementation Details}
The input RGB-color image is first transformed into the Hue-Saturation-Value (HSV) space.\ Since the Value (V) channel reflects the illumination and reflectance information, we only process this channel, and transform the processed image from the HSV space to RGB-color space, similar to~\cite{wvm2016,jiep2017}.\ In our experiments, we empirically set the parameters as $\alpha=0.001,\beta=0.0001,\gamma_{s}=1.5,\gamma_{t}=0.5$.\ We also compare with a Baseline of our STAR, in which we set $\gamma_{s}=\gamma_{t}=1$ in (\ref{e-star}).\ Due to considering local variance information, the EMLV filter (Eqn.~\ref{e-mlvf}) can reveal details and preserve structures better than the ETV filter (Figure~\ref{f-priors}).\ We will perform ablation study on these points in \S\ref{sec:validation}.\

\begin{figure*}
\centering
\begin{minipage}[t]{0.19\textwidth}
\begin{subfigure}[t]{1\linewidth}
\raisebox{-0.15cm}{\includegraphics[width=1\textwidth]{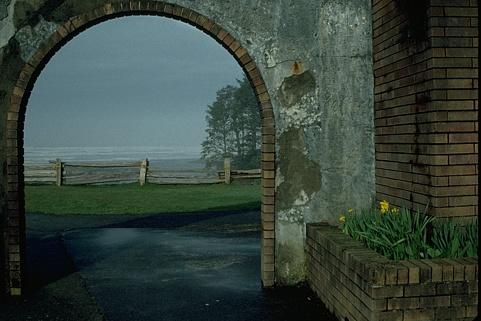}}
\centering{\scriptsize (a) Original Image}
\end{subfigure}
\end{minipage}
\begin{minipage}[t]{0.19\textwidth}
\begin{subfigure}[t]{1\linewidth}
\raisebox{-0.15cm}{\includegraphics[width=1\textwidth]{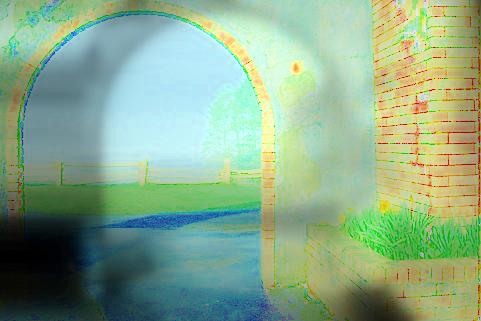}}
\centering{\scriptsize (b) Illumination by MSR~\cite{msrcr1997}}
\end{subfigure}
\begin{subfigure}[t]{1\linewidth}
\raisebox{-0.15cm}{\includegraphics[width=1\textwidth]{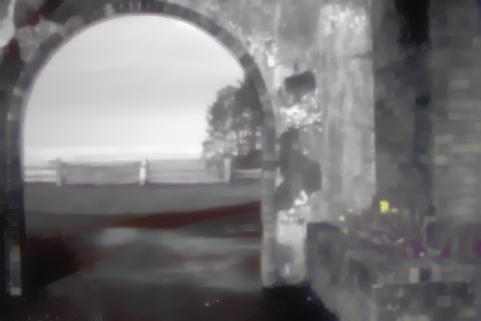}}
\centering{\scriptsize (f) Illumination by RRM~\cite{li2018structure}}
\end{subfigure}
\begin{subfigure}[t]{1\linewidth}
\raisebox{-0.15cm}{\includegraphics[width=1\textwidth]{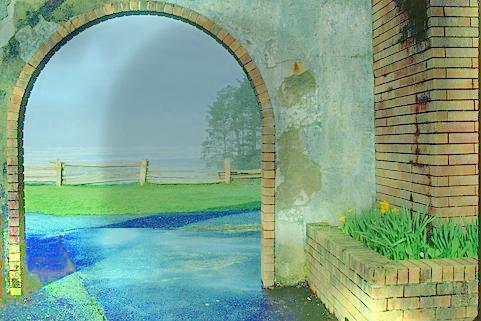}}
\centering{\scriptsize (j) Reflectance by MSR~\cite{msrcr1997}}
\end{subfigure}
\begin{subfigure}[t]{1\linewidth}
\raisebox{-0.15cm}{\includegraphics[width=1\textwidth]{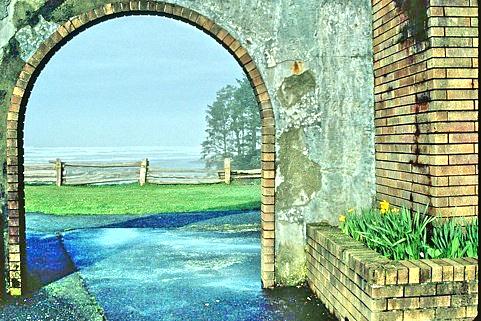}}
\centering{\scriptsize (n) Reflectance by RRM~\cite{li2018structure}}
\end{subfigure}
\end{minipage}
\begin{minipage}[t]{0.19\textwidth}
\begin{subfigure}[t]{1\linewidth}
\raisebox{-0.15cm}{\includegraphics[width=1\textwidth]{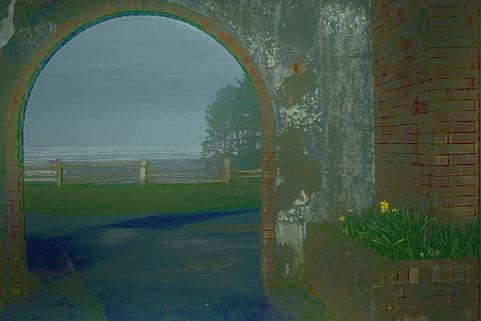}}
\centering{\scriptsize (c) Illumination by SIRE~\cite{srie2015}}
\end{subfigure}
\begin{subfigure}[t]{1\linewidth}
\raisebox{-0.15cm}{\includegraphics[width=1\textwidth]{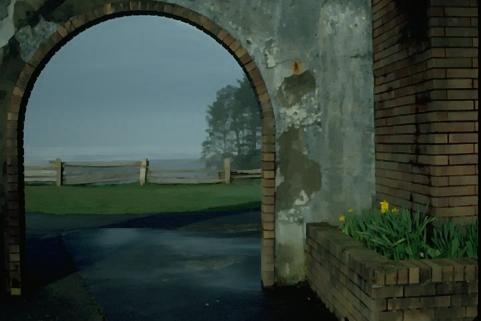}}
\centering{\scriptsize (g) Illumination by RDGAN~\cite{rdgan}}
\end{subfigure}
\begin{subfigure}[t]{1\linewidth}
\raisebox{-0.15cm}{\includegraphics[width=1\textwidth]{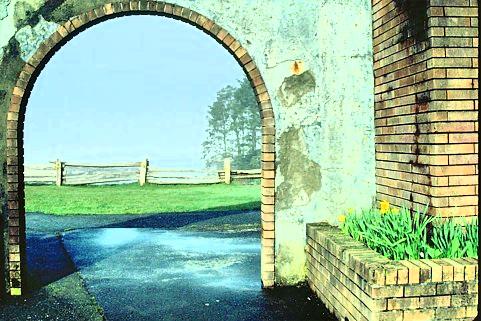}}
\centering{\scriptsize (k) Reflectance by SIRE~\cite{srie2015}}
\end{subfigure}
\begin{subfigure}[t]{1\linewidth}
\raisebox{-0.15cm}{\includegraphics[width=1\textwidth]{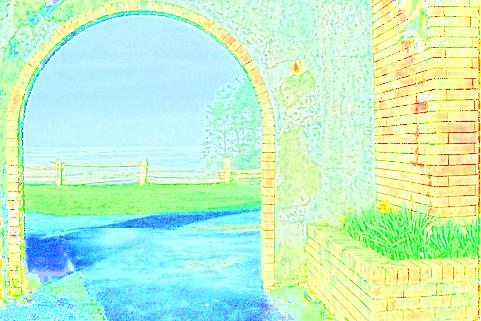}}
\centering{\scriptsize (o) Reflectance by RDGAN~\cite{rdgan}}
\end{subfigure}
\end{minipage}
\begin{minipage}[t]{0.19\textwidth}
\begin{subfigure}[t]{1\linewidth}
\raisebox{-0.15cm}{\includegraphics[width=1\textwidth]{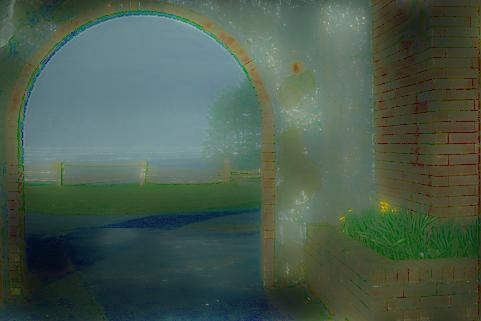}}
\centering{\scriptsize (d) Illumination by WVM~\cite{wvm2016}}
\end{subfigure}
\begin{subfigure}[t]{1\linewidth}
\raisebox{-0.15cm}{\includegraphics[width=1\textwidth]{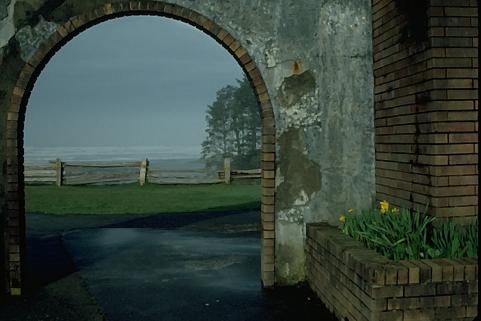}}
\centering{\scriptsize (h) Illumination by Baseline}
\end{subfigure}
\begin{subfigure}[t]{1\linewidth}
\raisebox{-0.15cm}{\includegraphics[width=1\textwidth]{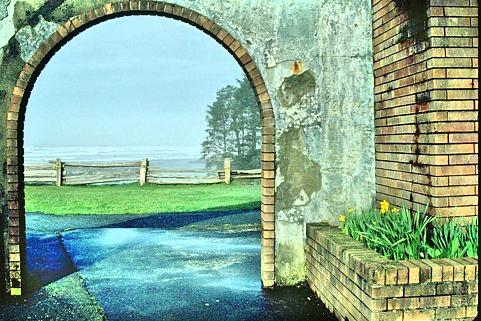}}
\centering{\scriptsize (l) Reflectance by WVM~\cite{wvm2016}}
\end{subfigure}
\begin{subfigure}[t]{1\linewidth}
\raisebox{-0.15cm}{\includegraphics[width=1\textwidth]{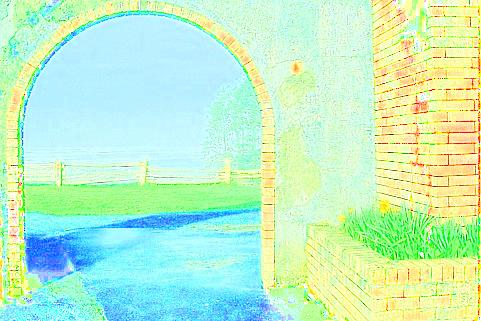}}
\centering{\scriptsize (p) Reflectance by Baseline}
\end{subfigure}
\end{minipage}
\begin{minipage}[t]{0.19\textwidth}
\begin{subfigure}[t]{1\linewidth}
\raisebox{-0.15cm}{\includegraphics[width=1\textwidth]{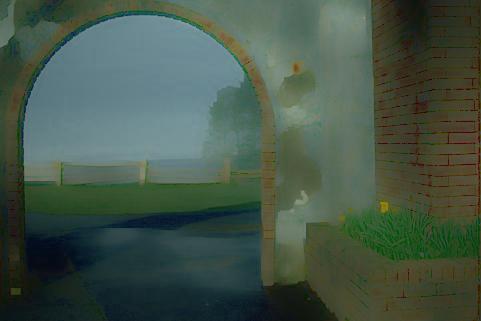}}
\centering{\scriptsize (e) Illumination by JieP~\cite{jiep2017}}
\end{subfigure}
\begin{subfigure}[t]{1\linewidth}
\raisebox{-0.15cm}{\includegraphics[width=1\textwidth]{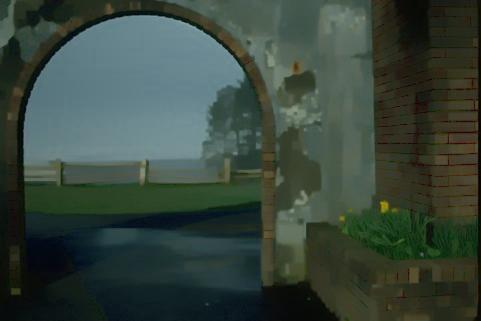}}
\centering{\scriptsize (i) Illumination by our STAR}
\end{subfigure}
\begin{subfigure}[t]{1\linewidth}
\raisebox{-0.15cm}{\includegraphics[width=1\textwidth]{F-Retinex/rs_1_R_JieP.jpg}}
\centering{\scriptsize (m) Reflectance by JieP~\cite{jiep2017}}
\end{subfigure}
\begin{subfigure}[t]{1\linewidth}
\raisebox{-0.15cm}{\includegraphics[width=1\textwidth]{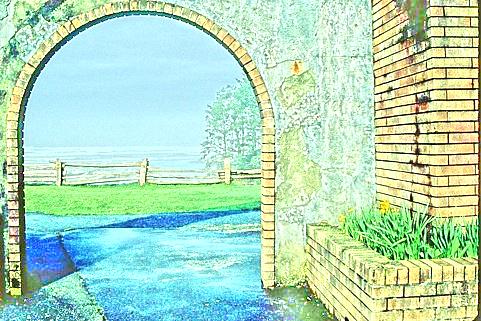}}
\centering{\scriptsize (q) Reflectance by our STAR}
\end{subfigure}
\end{minipage}
\vspace{-1mm}
\caption{\linespread{1}\selectfont{Comparisons of illumination and reflectance components by different Retinex decomposition methods on the image ``2'' from the 35 low-light images collected from~\cite{srie2015,wvm2016,lime2017,jiep2017}.}}
\label{F1-retinex}
\end{figure*}

\begin{figure*}
\centering
\begin{minipage}[t]{0.19\textwidth}
\begin{subfigure}[t]{1\linewidth}
\raisebox{-0.15cm}{\includegraphics[width=1\textwidth]{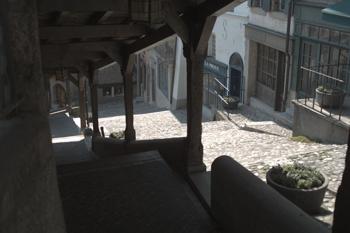}}
\centering{\scriptsize (a) Original Image}
\end{subfigure}
\end{minipage}
\begin{minipage}[t]{0.19\textwidth}
\begin{subfigure}[t]{1\linewidth}
\raisebox{-0.15cm}{\includegraphics[width=1\textwidth]{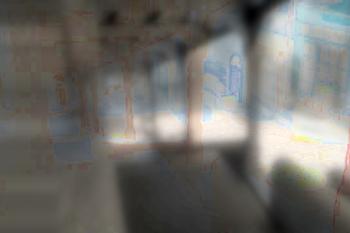}}
\centering{\scriptsize (b) Illumination by MSR~\cite{msrcr1997}}
\end{subfigure}
\begin{subfigure}[t]{1\linewidth}
\raisebox{-0.15cm}{\includegraphics[width=1\textwidth]{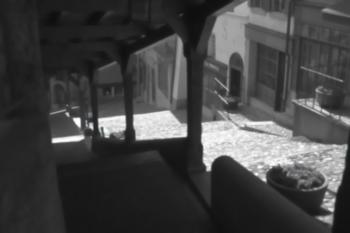}}
\centering{\scriptsize (f) Illumination by RRM~\cite{li2018structure}}
\end{subfigure}
\begin{subfigure}[t]{1\linewidth}
\raisebox{-0.15cm}{\includegraphics[width=1\textwidth]{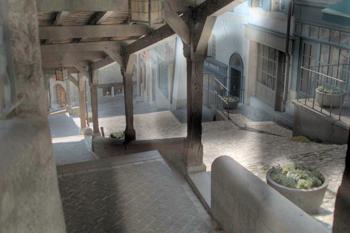}}
\centering{\scriptsize (j) Reflectance by MSR~\cite{msrcr1997}}
\end{subfigure}
\begin{subfigure}[t]{1\linewidth}
\raisebox{-0.15cm}{\includegraphics[width=1\textwidth]{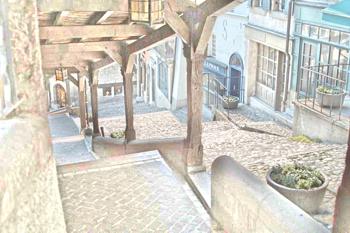}}
\centering{\scriptsize (n) Reflectance by RRM~\cite{li2018structure}}
\end{subfigure}
\end{minipage}
\begin{minipage}[t]{0.19\textwidth}
\begin{subfigure}[t]{1\linewidth}
\raisebox{-0.15cm}{\includegraphics[width=1\textwidth]{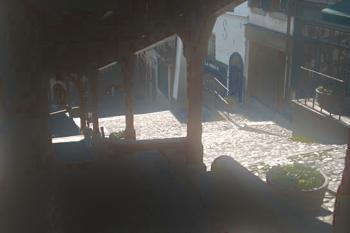}}
\centering{\scriptsize (c) Illumination by SIRE~\cite{srie2015}}
\end{subfigure}
\begin{subfigure}[t]{1\linewidth}
\raisebox{-0.15cm}{\includegraphics[width=1\textwidth]{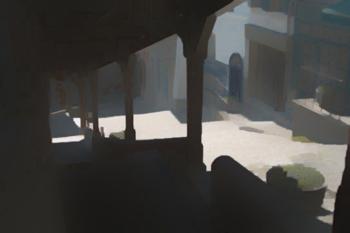}}
\centering{\scriptsize (g) Illumination by RDGAN~\cite{rdgan}}
\end{subfigure}
\begin{subfigure}[t]{1\linewidth}
\raisebox{-0.15cm}{\includegraphics[width=1\textwidth]{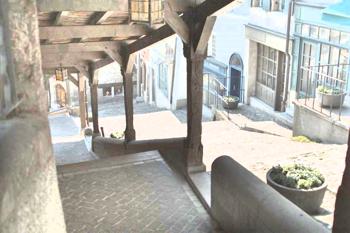}}
\centering{\scriptsize (k) Reflectance by SIRE~\cite{srie2015}}
\end{subfigure}
\begin{subfigure}[t]{1\linewidth}
\raisebox{-0.15cm}{\includegraphics[width=1\textwidth]{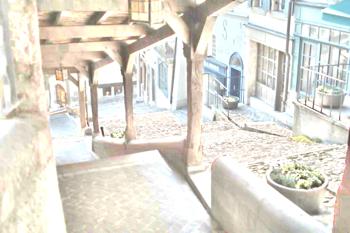}}
\centering{\scriptsize (o) Reflectance by RDGAN~\cite{rdgan}}
\end{subfigure}
\end{minipage}
\begin{minipage}[t]{0.19\textwidth}
\begin{subfigure}[t]{1\linewidth}
\raisebox{-0.15cm}{\includegraphics[width=1\textwidth]{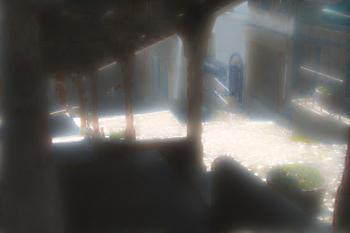}}
\centering{\scriptsize (d) Illumination by WVM~\cite{wvm2016}}
\end{subfigure}
\begin{subfigure}[t]{1\linewidth}
\raisebox{-0.15cm}{\includegraphics[width=1\textwidth]{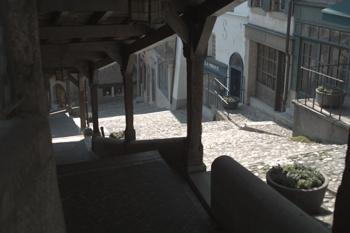}}
\centering{\scriptsize (h) Illumination by Baseline}
\end{subfigure}
\begin{subfigure}[t]{1\linewidth}
\raisebox{-0.15cm}{\includegraphics[width=1\textwidth]{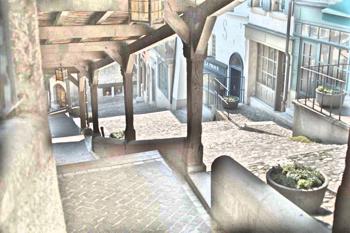}}
\centering{\scriptsize (l) Reflectance by WVM~\cite{wvm2016}}
\end{subfigure}
\begin{subfigure}[t]{1\linewidth}
\raisebox{-0.15cm}{\includegraphics[width=1\textwidth]{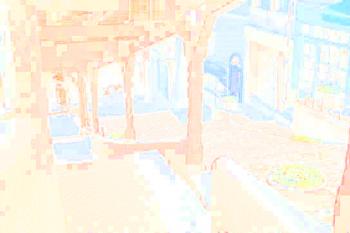}}
\centering{\scriptsize (p) Reflectance by Baseline}
\end{subfigure}
\end{minipage}
\begin{minipage}[t]{0.19\textwidth}
\begin{subfigure}[t]{1\linewidth}
\raisebox{-0.15cm}{\includegraphics[width=1\textwidth]{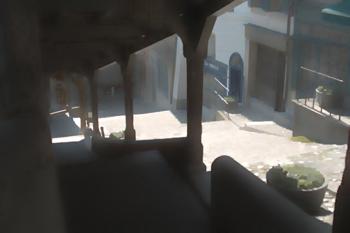}}
\centering{\scriptsize (e) Illumination by JieP~\cite{jiep2017}}
\end{subfigure}
\begin{subfigure}[t]{1\linewidth}
\raisebox{-0.15cm}{\includegraphics[width=1\textwidth]{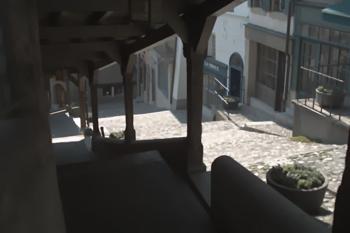}}
\centering{\scriptsize (i) Illumination by our STAR}
\end{subfigure}
\begin{subfigure}[t]{1\linewidth}
\raisebox{-0.15cm}{\includegraphics[width=1\textwidth]{F-Retinex/rs_7_R_JieP.jpg}}
\centering{\scriptsize (m) Reflectance by JieP~\cite{jiep2017}}
\end{subfigure}
\begin{subfigure}[t]{1\linewidth}
\raisebox{-0.15cm}{\includegraphics[width=1\textwidth]{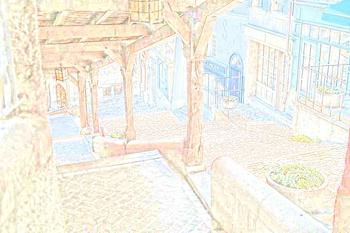}}
\centering{\scriptsize (q) Reflectance by our STAR}
\end{subfigure}
\end{minipage}
\vspace{-1mm}
\caption{\linespread{1}\selectfont{Comparisons of illumination and reflectance components by different Retinex decomposition methods on the image ``15'' from the 35 low-light images collected from~\cite{srie2015,wvm2016,lime2017,jiep2017}.}}
\label{F2-retinex}
\end{figure*}

%-------------------------------------------------------------------------
\subsection{Retinex Decomposition}
\label{sec:decom}
The Retinex decomposition includes illumination and reflectance estimation.\ Accurate illumination estimation should not distort the structure, while being spatially smooth.\ Meanwhile, accurate reflectance should reveal the details of the observed scene.\ The ground truth{s} for the illumination and reflectance components are difficult to generate, and hence quantitative evaluation of existing Retinex decomposition methods is very difficult until now.\

To evaluate the effectiveness of the proposed STAR model, we perform qualitative comparisons on both illumination and reflectance estimation with the Baseline, the conventional Multi-scale Retinex (MSR)~\cite{msrcr1997}, and several state-of-the-art Retinex models, including Simultaneous Illumination and Reflectance Estimation (SIRE)~\cite{srie2015}, Weighted Variation Model (WVM)~\cite{wvm2016}, Joint intrinsic-extrinsic Prior (JieP) model~\cite{jiep2017}, Robust Retinex Method (RRM)~\cite{li2018structure}, and Retinex Decomposition based Generative Adversarial Network (RDGAN)~\cite{rdgan}.\ Similar to these methods, we perform Retinex decomposition on the V channel of the HSV space, and transform the decomposed components back to the RGB space.\ Some visual results on two common test images in the 35 low-light images collected from~\cite{srie2015,wvm2016,lime2017,jiep2017} are shown in Figures~\ref{F1-retinex} and~\ref{F2-retinex}.\ It can be seen that, in the proposed STAR model, the structure awareness scheme enforces piece-wise smoothness, while the texture awareness scheme preserves details across the image.\ As can be seen in Figures~\ref{F1-retinex} and \ref{F2-retinex} (h), (i), (e)-(g), the proposed STAR method preserves better the structure of the three black regions on the white car, and reveals more details of the texture on the wall, than the Baseline and the other methods such as WVM~\cite{wvm2016}, JieP~\cite{jiep2017}, and RRM~\cite{li2018structure}.\ More comparisons on Retinex decomposition are provided in the \textsl{Supplementary File}.

\noindent
\textbf{Comparison on speed}.\ In Table~\ref{T-Time}, we also compare the computational time of different Retinex image decomposition methods on a $960\times720$ RGB image.\ We observe that our STAR is faster than WVM~\cite{wvm2016} and RRM~\cite{li2018structure}, but slower than JieP~\cite{jiep2017} and SIRE~\cite{srie2015}.\ Though not the fastest method, our STAR achieves better decomposition performance than the other methods such as JieP~\cite{jiep2017} and SIRE~\cite{srie2015}. 
\begin{table}[htp]
\begin{center}
\begin{tabular}{ccccc}
\Xhline{1pt}
\rowcolor[rgb]{ .9,  .9,  .9}
Method
&
MSR~\cite{msrcr1997}
&
SIRE~\cite{srie2015}
&
WVM~\cite{wvm2016}
&
JieP~\cite{jiep2017}
\\
\hline
Time
& 3.16 & 2.83 & 58.48 & 16.40
\\
\hline
\rowcolor[rgb]{ .9,  .9,  .9}
Method
&
RRM~\cite{li2018structure}
&
RDGAN~\cite{rdgan}
&
Baseline
&
STAR
\\
\hline
Time
& 107.72 & 26.07 & 4.02 & 23.52
\\
\hline
\end{tabular} 
\vspace{-1mm}
\caption{Comparison on computational times (in seconds) of different Retinex image decomposition methods.\ The time are computed by averaging the 10 running times of these methods on a RGB image of size $960\times720$.}
\label{T-Time}
\vspace{-3mm}
\end{center}
\end{table}

\begin{figure*}
\centering
\begin{minipage}[t]{0.32\textwidth}
\begin{subfigure}[b]{1\linewidth}
\raisebox{-0.15cm}{\includegraphics[width=1\textwidth]{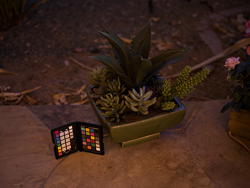}}
\centering{\scriptsize (a) Input }
\end{subfigure}
\begin{subfigure}[b]{1\linewidth}
\raisebox{-0.15cm}{\includegraphics[width=1\textwidth]{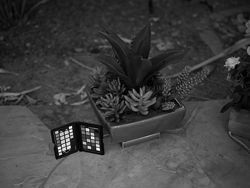}}
\centering{\scriptsize (d) V channel of (a) }
\end{subfigure}
\end{minipage}
\begin{minipage}[t]{0.32\textwidth}
\begin{subfigure}[b]{1\linewidth}
\raisebox{-0.15cm}{\includegraphics[width=1\textwidth]{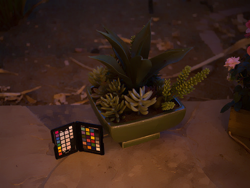}}
\centering{\scriptsize (b) Illumination by STAR-ETV}
\end{subfigure}
\begin{subfigure}[b]{1\linewidth}
\raisebox{-0.15cm}{\includegraphics[width=1\textwidth]{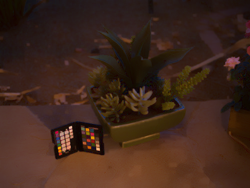}}
\centering{\scriptsize (e) Illumination by STAR-EMLV}
\end{subfigure}
\end{minipage}
\begin{minipage}[t]{0.32\textwidth}
\begin{subfigure}[b]{1\linewidth}
\raisebox{-0.15cm}{\includegraphics[width=1\textwidth]{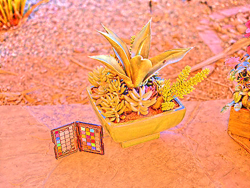}}
\centering{\scriptsize (c) Reflectance by STAR-ETV}
\end{subfigure}
\begin{subfigure}[b]{1\linewidth}
\raisebox{-0.15cm}{\includegraphics[width=1\textwidth]{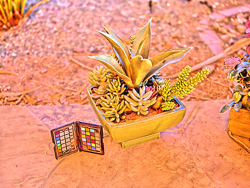}}
\centering{\scriptsize (f) Reflectance by STAR-EMLV}
\end{subfigure}
\end{minipage}
\vspace{-1mm}
\caption{\linespread{1}\selectfont{
Comparison of illumination and reflectance components by the proposed STAR model with ETV or EMLV weighting schemes.\ For better comparisons, we illustrate the components in the RGB channels instead of V channel.}}
\vspace{-1mm}
\label{F-Validation-1}
\end{figure*}

\begin{figure*}
\centering
\begin{minipage}[t]{0.24\textwidth}
\begin{subfigure}[b]{1\linewidth}
\raisebox{-0.15cm}{\includegraphics[width=1\textwidth]{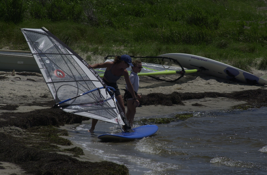}}
\centering{\scriptsize (a) Input }
\end{subfigure}
\begin{subfigure}[b]{1\linewidth}
\raisebox{-0.15cm}{\includegraphics[width=1\textwidth]{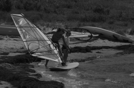}}
\centering{\scriptsize (e) V channel of (a) }
\end{subfigure}
\end{minipage}
\begin{minipage}[t]{0.24\textwidth}
\begin{subfigure}[b]{1\linewidth}
\raisebox{-0.15cm}{\includegraphics[width=1\textwidth]{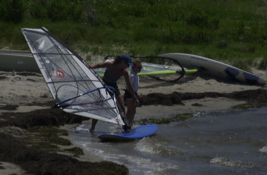}}
\centering{\scriptsize (b) Illumination by STAR \textsl{w/o} $\bm{S}$}
\end{subfigure}
\begin{subfigure}[b]{1\linewidth}
\raisebox{-0.15cm}{\includegraphics[width=1\textwidth]{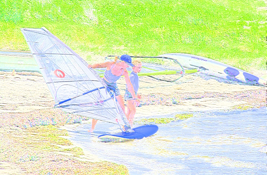}}
\centering{\scriptsize (f) Reflectance by STAR \textsl{w/o} $\bm{S}$}
\end{subfigure}
\end{minipage}
\begin{minipage}[t]{0.24\textwidth}
\begin{subfigure}[b]{1\linewidth}
\raisebox{-0.15cm}{\includegraphics[width=1\textwidth]{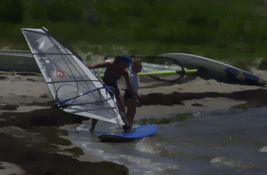}}
\centering{\scriptsize (c) Illumination by STAR \textsl{w/o} $\bm{T}$}
\end{subfigure}
\begin{subfigure}[b]{1\linewidth}
\raisebox{-0.15cm}{\includegraphics[width=1\textwidth]{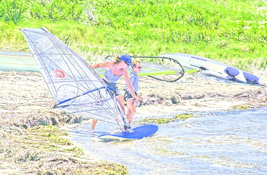}}
\centering{\scriptsize (g) Reflectance by STAR \textsl{w/o} $\bm{T}$}
\end{subfigure}
\end{minipage}
\begin{minipage}[t]{0.24\textwidth}
\begin{subfigure}[b]{1\linewidth}
\raisebox{-0.15cm}{\includegraphics[width=1\textwidth]{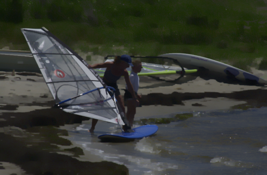}}
\centering{\scriptsize (d) Illumination by STAR}
\end{subfigure}
\begin{subfigure}[b]{1\linewidth}
\raisebox{-0.15cm}{\includegraphics[width=1\textwidth]{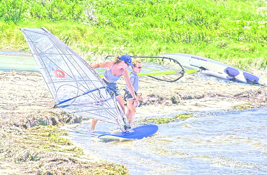}}
\centering{\scriptsize (h) Reflectance by STAR}
\end{subfigure}
\end{minipage}
\vspace{-1mm}
\caption{\linespread{1}\selectfont{
Ablation study of structure or texture map on the illumination and reflectance decomposition performance of the proposed STAR model.\ For better comparisons, we illustrate the components in the RGB channels instead of V channel.}}
\vspace{-1mm}
\label{F-Validation-2}
\end{figure*}

\begin{figure*}
\centering
\begin{minipage}[t]{0.24\textwidth}
\begin{subfigure}[b]{1\linewidth}
\raisebox{-0.15cm}{\includegraphics[width=1\textwidth]{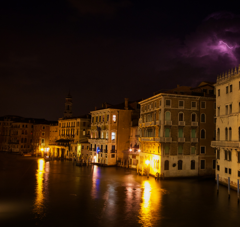}}
\centering{\scriptsize (a) Input }
\end{subfigure}
\begin{subfigure}[b]{1\linewidth}
\raisebox{-0.15cm}{\includegraphics[width=1\textwidth]{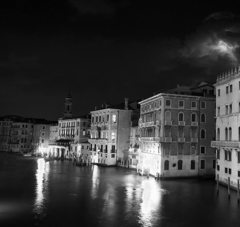}}
\centering{\scriptsize (e) V channel of (a) }
\end{subfigure}
\end{minipage}
\begin{minipage}[t]{0.24\textwidth}
\begin{subfigure}[b]{1\linewidth}
\raisebox{-0.15cm}{\includegraphics[width=1\textwidth]{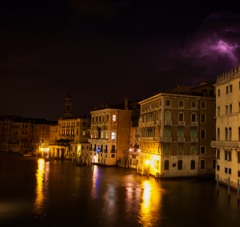}}
\centering{\scriptsize (b) Illumination with $\gamma_{s}=0.5,\gamma_{t}=1.5$}
\end{subfigure}
\begin{subfigure}[b]{1\linewidth}
\raisebox{-0.15cm}{\includegraphics[width=1\textwidth]{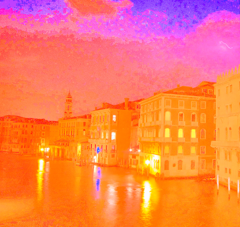}}
\centering{\scriptsize (f) Reflectance with $\gamma_{s}=0.5,\gamma_{t}=1.5$}
\end{subfigure}
\end{minipage}
\begin{minipage}[t]{0.24\textwidth}
\begin{subfigure}[b]{1\linewidth}
\raisebox{-0.15cm}{\includegraphics[width=1\textwidth]{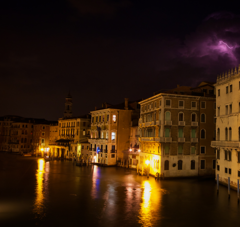}}
\centering{\scriptsize (c) Illumination with $\gamma_{s}=1,\gamma_{t}=1$}
\end{subfigure}
\begin{subfigure}[b]{1\linewidth}
\raisebox{-0.15cm}{\includegraphics[width=1\textwidth]{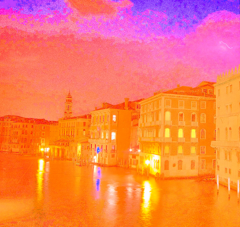}}
\centering{\scriptsize (g) Reflectance with $\gamma_{s}=1,\gamma_{t}=1$}
\end{subfigure}
\end{minipage}
\begin{minipage}[t]{0.24\textwidth}
\begin{subfigure}[b]{1\linewidth}
\raisebox{-0.15cm}{\includegraphics[width=1\textwidth]{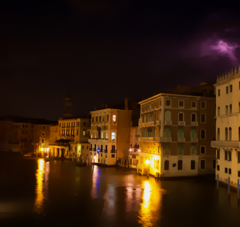}}
\centering{\scriptsize (d) Illumination with $\gamma_{s}=1.5,\gamma_{t}=0.5$}
\end{subfigure}
\begin{subfigure}[b]{1\linewidth}
\raisebox{-0.15cm}{\includegraphics[width=1\textwidth]{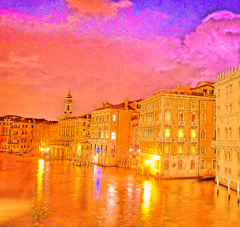}}
\centering{\scriptsize (h) Reflectance with $\gamma_{s}=1.5,\gamma_{t}=0.5$}
\end{subfigure}
\end{minipage}
\vspace{-1mm}
\caption{\linespread{1}\selectfont{
Ablation study of the parameters ($\gamma_{s},\gamma_{t}$) on the illumination and reflectance decomposition performance of the proposed STAR model.\ For better comparisons, we illustrate the components in the RGB channels instead of V channel.}}
\label{F-Validation-3}
\end{figure*}

\begin{figure*}
\centering
\begin{minipage}[t]{0.24\textwidth}
\begin{subfigure}[b]{1\linewidth}
\raisebox{-0.15cm}{\includegraphics[width=1\textwidth]{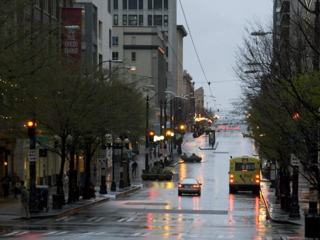}}
\centering{\scriptsize (a) Input }
\end{subfigure}
\begin{subfigure}[b]{1\linewidth}
\raisebox{-0.15cm}{\includegraphics[width=1\textwidth]{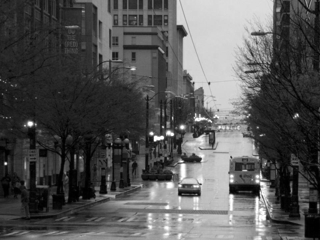}}
\centering{\scriptsize (e) V channel of (a) }
\end{subfigure}
\end{minipage}
\begin{minipage}[t]{0.24\textwidth}
\begin{subfigure}[b]{1\linewidth}
\raisebox{-0.15cm}{\includegraphics[width=1\textwidth]{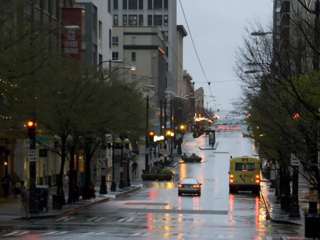}}
\centering{\scriptsize (b) Illumination with $L=1$}
\end{subfigure}
\begin{subfigure}[b]{1\linewidth}
\raisebox{-0.15cm}{\includegraphics[width=1\textwidth]{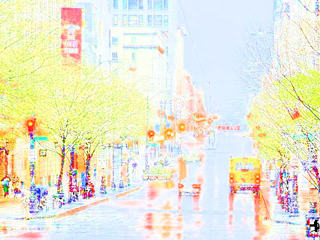}}
\centering{\scriptsize (f) Reflectance with $L=1$}
\end{subfigure}
\end{minipage}
\begin{minipage}[t]{0.24\textwidth}
\begin{subfigure}[b]{1\linewidth}
\raisebox{-0.15cm}{\includegraphics[width=1\textwidth]{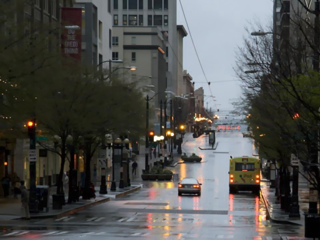}}
\centering{\scriptsize (c) Illumination with $L=2$}
\end{subfigure}
\begin{subfigure}[b]{1\linewidth}
\raisebox{-0.15cm}{\includegraphics[width=1\textwidth]{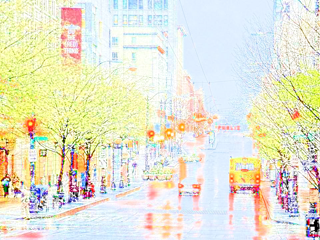}}
\centering{\scriptsize (g) Reflectance with $L=2$}
\end{subfigure}
\end{minipage}
\begin{minipage}[t]{0.24\textwidth}
\begin{subfigure}[b]{1\linewidth}
\raisebox{-0.15cm}{\includegraphics[width=1\textwidth]{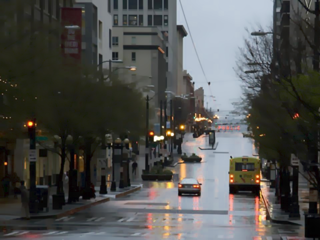}}
\centering{\scriptsize (d) Illumination with $L=4$}
\end{subfigure}
\begin{subfigure}[b]{1\linewidth}
\raisebox{-0.15cm}{\includegraphics[width=1\textwidth]{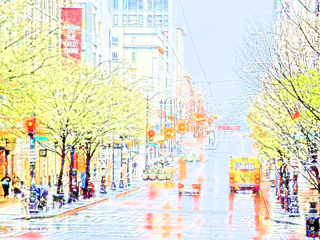}}
\centering{\scriptsize (h) Reflectance with $L=4$}
\end{subfigure}
\end{minipage}
\vspace{-1mm}
\caption{\linespread{1}\selectfont{
Ablation study of updating iterations $L$ on the illumination and reflectance decomposition performance of the proposed STAR model.\ For better comparisons, we illustrate the components in the RGB channels instead of V channel.}}
\label{F-Validation-4}
\end{figure*}

%-------------------------
\subsection{Validation of the Proposed STAR Model}
\label{sec:validation}
Here, we conduct a detailed examination of our STAR for Retinex decomposition.\ We assess 1) the choice of the weighting scheme (ETV or EMLV) on our STAR; 2) the importance of structure and texture awareness to our STAR; 3) the influence of the parameters $\gamma_{s},\gamma_{t}$ on our STAR; 4) how to determine the parameters $\alpha$ and $\beta$ in our STAR? 5) the necessity of updating structure $\bm{S}$ and texture $\bm{T}$ to our STAR.

\noindent
\textbf{1. The influence of the weighting scheme (ETV or EMLV) on our STAR}.\ To study the the weighting scheme (ETV or EMLV) on our STAR, we employ the ETV filter~(\ref{e-tvf}) and set $\bm{S}_{0}=1/(|\nabla\bm{I}_{0}|+\varepsilon)$ and $\bm{T}_{0}=1/(|\nabla\bm{O}_{0}|+\varepsilon)$ in (\ref{e-star}) and update them as Algorithm 2 describes, and thus have another STAR model: \textsl{STAR-ETV}.\ The default STAR model can be termed as \textsl{STAR-EMLV}.\ From Figure~\ref{F-Validation-1}, one can see that, the \textsl{STAR-ETV} model tends to provide little structure in illumination, while losing texture information in reflectance.\ By employing EMLV filter as the weighting matrix, the proposed STAR (STAR-EMLV) method maintains the structure and texture better than the \textsl{STAR-ETV} model.

\noindent
\textbf{2. Is structure and texture awareness important?} To answer this question, we set $\bm{S}_{0}=1/(|\nabla\bm{I}_{0}|+\varepsilon)$ or $\bm{T}_{0}=1/(|\nabla\bm{O}_{0}|+\varepsilon)$ in (\ref{e-star}) and update them as Algorithm 2 describes, and thus have two baselines: \textsl{STAR w/o Structure} and \textsl{STAR w/o Texture}.\ Note that if we set $\bm{S}_{0}$ or $\bm{T}_{0}$ in (\ref{e-star}) as comfortable identity matrix, the performance of the corresponding STAR model is very bad.\ From Figure~\ref{F-Validation-2}, one can see that, \textsl{STAR w/o Structure} tends to provide little structural information in illumination, while \textsl{STAR w/o Texture} influence little in illumination and reflectance.\ By considering both, the proposed STAR decompose the structure/texture components accurately.

\noindent
\textbf{3. How do the parameters $\gamma_{s}$ and $\gamma_{t}$ influence STAR?}
The $\gamma_{s},\gamma_{t}$ are key parameters for the structure and texture awareness of STAR.\ In Figure~\ref{F-Validation-3}, one can see that STAR with $\gamma_{s}=1.5,\gamma_{t}=0.5$ ((d) and (h)) produces reasonable results, STAR with $\gamma_{s}=1,\gamma_{t}=1$ ((c) and (g)) can barely distinguish the illumination and reflectance, while STAR with $\gamma_{s}=0.5,\gamma_{t}=1.5$ ((b) and (f)) confuses illumination and reflectance to a great extent.\ Since we regularize more on $\bm{I}$ ($\alpha=0.001,\beta=0.0001$), $\bm{I}$ and $\bm{R}$ in (f) are not exactly the same as $\bm{R}$ and $\bm{I}$ in (b), respectively.

\begin{figure*}
\centering
\begin{minipage}[t]{0.19\textwidth}
\begin{subfigure}[t]{1\linewidth}
\raisebox{-0.15cm}{\includegraphics[width=1\textwidth]{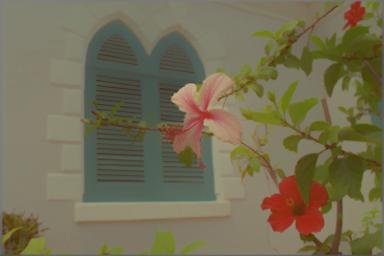}}
\centering{\scriptsize $\bm{I}, \alpha=10^{-1},\beta=1$}
\end{subfigure}
\begin{subfigure}[t]{1\linewidth}
\raisebox{-0.15cm}{\includegraphics[width=1\textwidth]{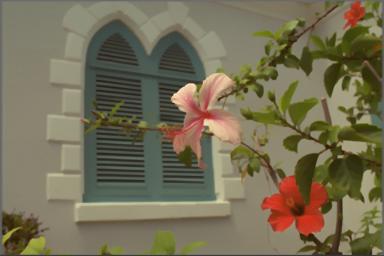}}
\centering{\scriptsize $\bm{I}, \alpha=10^{-2},\beta=1$}
\end{subfigure}
\begin{subfigure}[t]{1\linewidth}
\raisebox{-0.15cm}{\includegraphics[width=1\textwidth]{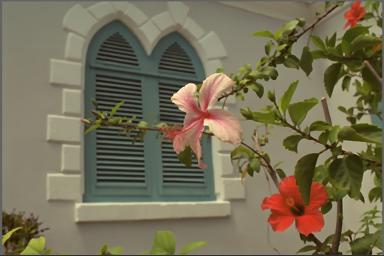}}
\centering{\scriptsize $\bm{I}, \alpha=10^{-3},\beta=1$}
\end{subfigure}
\begin{subfigure}[t]{1\linewidth}
\raisebox{-0.15cm}{\includegraphics[width=1\textwidth]{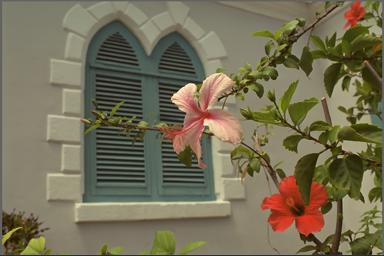}}
\centering{\scriptsize $\bm{I}, \alpha=10^{-4},\beta=1$}
\end{subfigure}
\begin{subfigure}[t]{1\linewidth}
\raisebox{-0.15cm}{\includegraphics[width=1\textwidth]{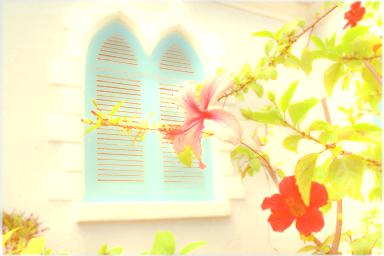}}
\centering{\scriptsize $\bm{R}, \alpha=10^{-1},\beta=1$}
\end{subfigure}
\begin{subfigure}[t]{1\linewidth}
\raisebox{-0.15cm}{\includegraphics[width=1\textwidth]{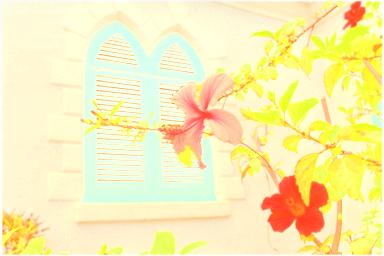}}
\centering{\scriptsize $\bm{R}, \alpha=10^{-2},\beta=1$}
\end{subfigure}
\begin{subfigure}[t]{1\linewidth}
\raisebox{-0.15cm}{\includegraphics[width=1\textwidth]{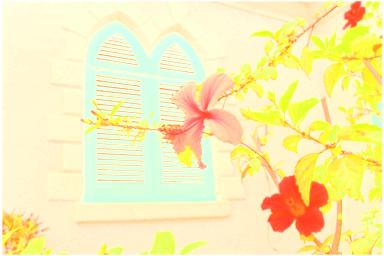}}
\centering{\scriptsize $\bm{R}, \alpha=10^{-3},\beta=1$}
\end{subfigure}
\begin{subfigure}[t]{1\linewidth}
\raisebox{-0.15cm}{\includegraphics[width=1\textwidth]{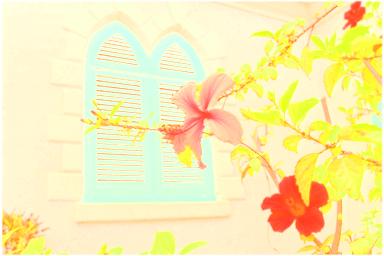}}
\centering{\scriptsize $\bm{R}, \alpha=10^{-4},\beta=1$}
\end{subfigure}
\end{minipage}
\begin{minipage}[t]{0.19\textwidth}
\begin{subfigure}[t]{1\linewidth}
\raisebox{-0.15cm}{\includegraphics[width=1\textwidth]{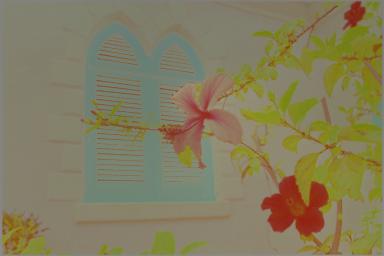}}
\centering{\scriptsize $\bm{I}, \alpha=10^{-1},\beta=10^{-1}$}
\end{subfigure}
\begin{subfigure}[t]{1\linewidth}
\raisebox{-0.15cm}{\includegraphics[width=1\textwidth]{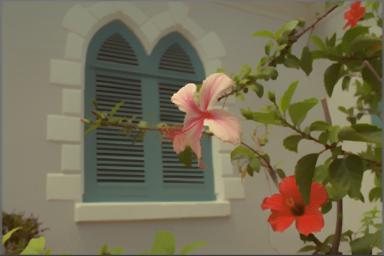}}
\centering{\scriptsize $\bm{I}, \alpha=10^{-2},\beta=10^{-1}$}
\end{subfigure}
\begin{subfigure}[t]{1\linewidth}
\raisebox{-0.15cm}{\includegraphics[width=1\textwidth]{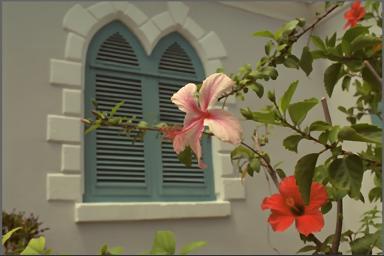}}
\centering{\scriptsize $\bm{I}, \alpha=10^{-3},\beta=10^{-1}$}
\end{subfigure}
\begin{subfigure}[t]{1\linewidth}
\raisebox{-0.15cm}{\includegraphics[width=1\textwidth]{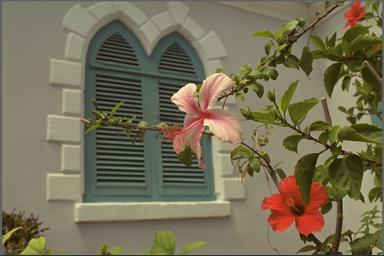}}
\centering{\scriptsize $\bm{I}, \alpha=10^{-4},\beta=10^{-1}$}
\end{subfigure}
\begin{subfigure}[t]{1\linewidth}
\raisebox{-0.15cm}{\includegraphics[width=1\textwidth]{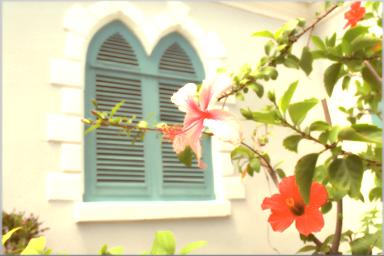}}
\centering{\scriptsize $\bm{R}, \alpha=10^{-1},\beta=10^{-1}$}
\end{subfigure}
\begin{subfigure}[t]{1\linewidth}
\raisebox{-0.15cm}{\includegraphics[width=1\textwidth]{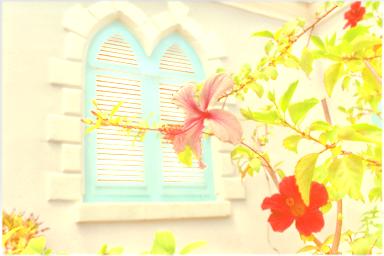}}
\centering{\scriptsize $\bm{R}, \alpha=10^{-2},\beta=10^{-1}$}
\end{subfigure}
\begin{subfigure}[t]{1\linewidth}
\raisebox{-0.15cm}{\includegraphics[width=1\textwidth]{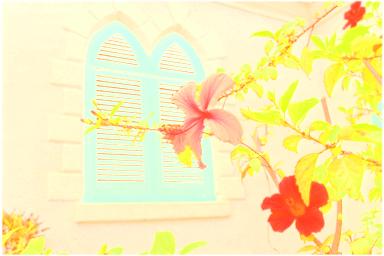}}
\centering{\scriptsize $\bm{R}, \alpha=10^{-3},\beta=10^{-1}$}
\end{subfigure}
\begin{subfigure}[t]{1\linewidth}
\raisebox{-0.15cm}{\includegraphics[width=1\textwidth]{F-AlphaBeta-2/0_R_RGB_STAR_alpha0001_beta01_pI15_pR05.jpg}}
\centering{\scriptsize $\bm{R}, \alpha=10^{-4},\beta=10^{-1}$}
\end{subfigure}
\end{minipage}
\begin{minipage}[t]{0.19\textwidth}
\begin{subfigure}[t]{1\linewidth}
\raisebox{-0.15cm}{\includegraphics[width=1\textwidth]{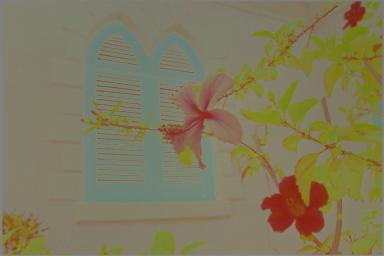}}
\centering{\scriptsize $\bm{I}, \alpha=10^{-1},\beta=10^{-2}$}
\end{subfigure}
\begin{subfigure}[t]{1\linewidth}
\raisebox{-0.15cm}{\includegraphics[width=1\textwidth]{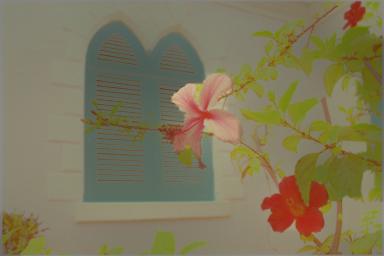}}
\centering{\scriptsize $\bm{I}, \alpha=10^{-2},\beta=10^{-2}$}
\end{subfigure}
\begin{subfigure}[t]{1\linewidth}
\raisebox{-0.15cm}{\includegraphics[width=1\textwidth]{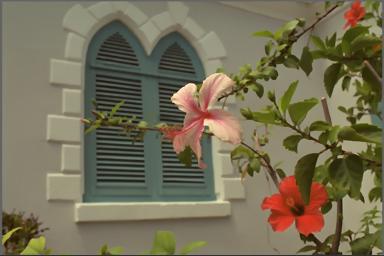}}
\centering{\scriptsize $\bm{I}, \alpha=10^{-3},\beta=10^{-2}$}
\end{subfigure}
\begin{subfigure}[t]{1\linewidth}
\raisebox{-0.15cm}{\includegraphics[width=1\textwidth]{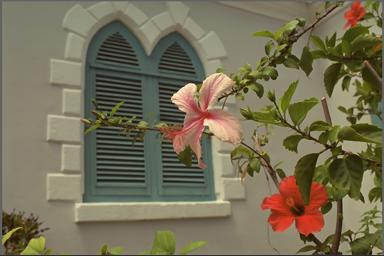}}
\centering{\scriptsize $\bm{I}, \alpha=10^{-4},\beta=10^{-2}$}
\end{subfigure}
\begin{subfigure}[t]{1\linewidth}
\raisebox{-0.15cm}{\includegraphics[width=1\textwidth]{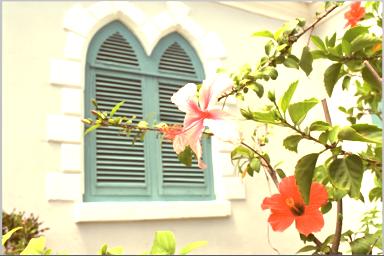}}
\centering{\scriptsize $\bm{R}, \alpha=10^{-1},\beta=10^{-2}$}
\end{subfigure}
\begin{subfigure}[t]{1\linewidth}
\raisebox{-0.15cm}{\includegraphics[width=1\textwidth]{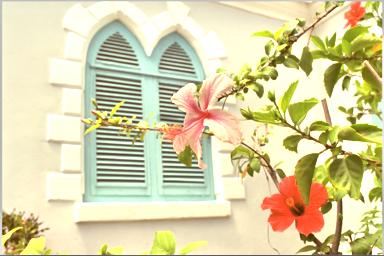}}
\centering{\scriptsize $\bm{R}, \alpha=10^{-2},\beta=10^{-2}$}
\end{subfigure}
\begin{subfigure}[t]{1\linewidth}
\raisebox{-0.15cm}{\includegraphics[width=1\textwidth]{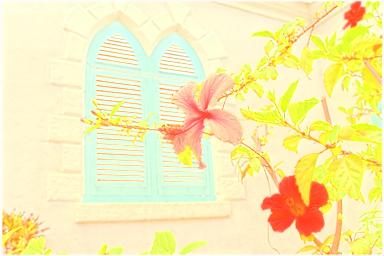}}
\centering{\scriptsize $\bm{R}, \alpha=10^{-3},\beta=10^{-2}$}
\end{subfigure}
\begin{subfigure}[t]{1\linewidth}
\raisebox{-0.15cm}{\includegraphics[width=1\textwidth]{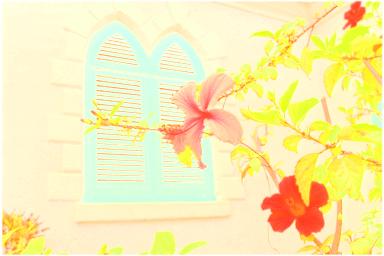}}
\centering{\scriptsize $\bm{R}, \alpha=10^{-4},\beta=10^{-2}$}
\end{subfigure}
\end{minipage}
\begin{minipage}[t]{0.19\textwidth}
\begin{subfigure}[t]{1\linewidth}
\raisebox{-0.15cm}{\includegraphics[width=1\textwidth]{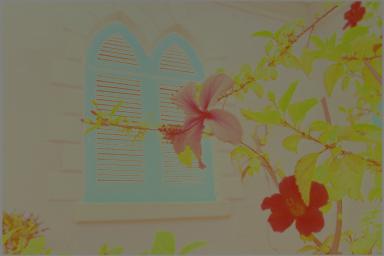}}
\centering{\scriptsize $\bm{I}, \alpha=10^{-1},\beta=10^{-3}$}
\end{subfigure}
\begin{subfigure}[t]{1\linewidth}
\raisebox{-0.15cm}{\includegraphics[width=1\textwidth]{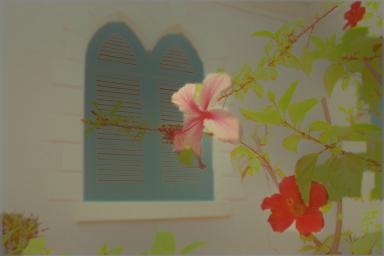}}
\centering{\scriptsize $\bm{I}, \alpha=10^{-2},\beta=10^{-3}$}
\end{subfigure}
\begin{subfigure}[t]{1\linewidth}
\raisebox{-0.15cm}{\includegraphics[width=1\textwidth]{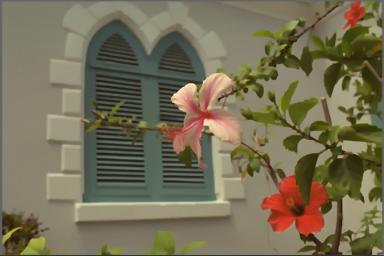}}
\centering{\scriptsize $\bm{I}, \alpha=10^{-3},\beta=10^{-3}$}
\end{subfigure}
\begin{subfigure}[t]{1\linewidth}
\raisebox{-0.15cm}{\includegraphics[width=1\textwidth]{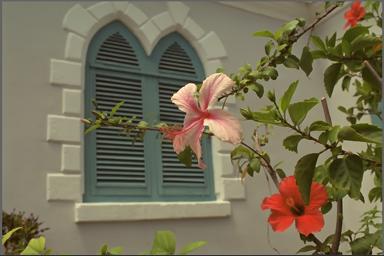}}
\centering{\scriptsize $\bm{I}, \alpha=10^{-4},\beta=10^{-3}$}
\end{subfigure}
\begin{subfigure}[t]{1\linewidth}
\raisebox{-0.15cm}{\includegraphics[width=1\textwidth]{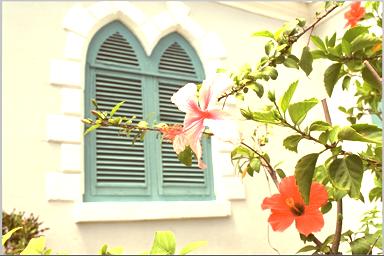}}
\centering{\scriptsize $\bm{R}, \alpha=10^{-1},\beta=10^{-3}$}
\end{subfigure}
\begin{subfigure}[t]{1\linewidth}
\raisebox{-0.15cm}{\includegraphics[width=1\textwidth]{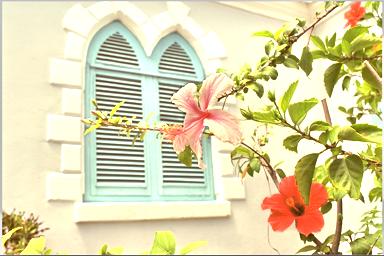}}
\centering{\scriptsize $\bm{R}, \alpha=10^{-2},\beta=10^{-3}$}
\end{subfigure}
\begin{subfigure}[t]{1\linewidth}
\raisebox{-0.15cm}{\includegraphics[width=1\textwidth]{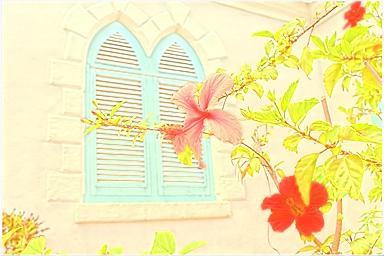}}
\centering{\scriptsize $\bm{R}, \alpha=10^{-3},\beta=10^{-3}$}
\end{subfigure}
\begin{subfigure}[t]{1\linewidth}
\raisebox{-0.15cm}{\includegraphics[width=1\textwidth]{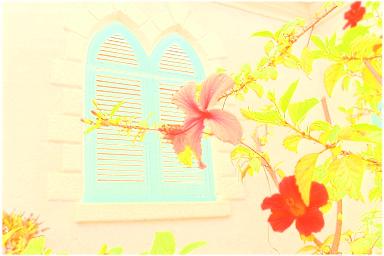}}
\centering{\scriptsize $\bm{R}, \alpha=10^{-4},\beta=10^{-3}$}
\end{subfigure}
\end{minipage}
\begin{minipage}[t]{0.19\textwidth}
\begin{subfigure}[t]{1\linewidth}
\raisebox{-0.15cm}{\includegraphics[width=1\textwidth]{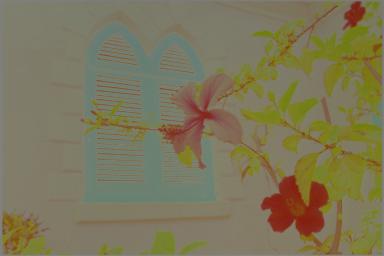}}
\centering{\scriptsize $\bm{I}, \alpha=10^{-1},\beta=10^{-4}$}
\end{subfigure}
\begin{subfigure}[t]{1\linewidth}
\raisebox{-0.15cm}{\includegraphics[width=1\textwidth]{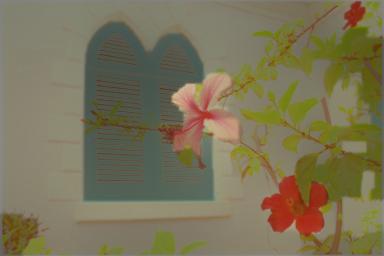}}
\centering{\scriptsize $\bm{I}, \alpha=10^{-2},\beta=10^{-4}$}
\end{subfigure}
\begin{subfigure}[t]{1\linewidth}
\raisebox{-0.15cm}{\includegraphics[width=1\textwidth]{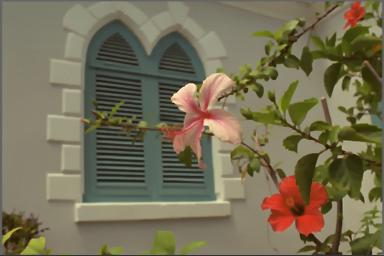}}
\centering{\scriptsize $\bm{I}, \alpha=10^{-3},\beta=10^{-4}$}
\end{subfigure}
\begin{subfigure}[t]{1\linewidth}
\raisebox{-0.15cm}{\includegraphics[width=1\textwidth]{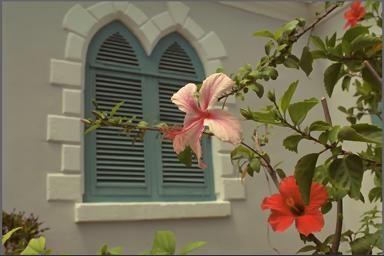}}
\centering{\scriptsize $\bm{I}, \alpha=10^{-4},\beta=10^{-4}$}
\end{subfigure}
\begin{subfigure}[t]{1\linewidth}
\raisebox{-0.15cm}{\includegraphics[width=1\textwidth]{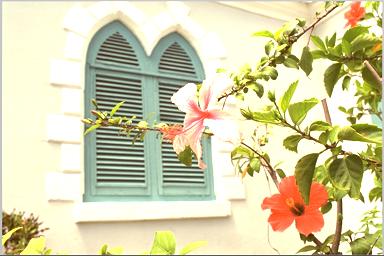}}
\centering{\scriptsize $\bm{R}, \alpha=10^{-1},\beta=10^{-4}$}
\end{subfigure}
\begin{subfigure}[t]{1\linewidth}
\raisebox{-0.15cm}{\includegraphics[width=1\textwidth]{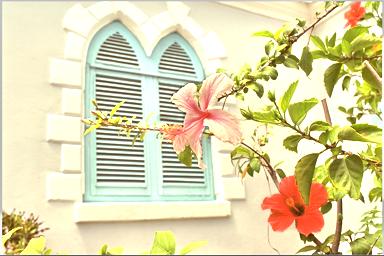}}
\centering{\scriptsize $\bm{R}, \alpha=10^{-2},\beta=10^{-4}$}
\end{subfigure}
\begin{subfigure}[t]{1\linewidth}
\raisebox{-0.15cm}{\includegraphics[width=1\textwidth]{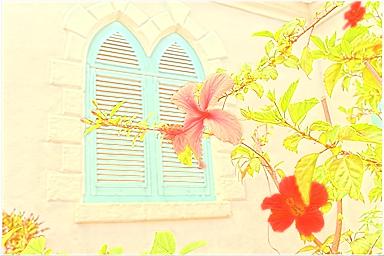}}
\centering{\scriptsize $\bm{R}, \alpha=10^{-3},\beta=10^{-4}$}
\end{subfigure}
\begin{subfigure}[t]{1\linewidth}
\raisebox{-0.15cm}{\includegraphics[width=1\textwidth]{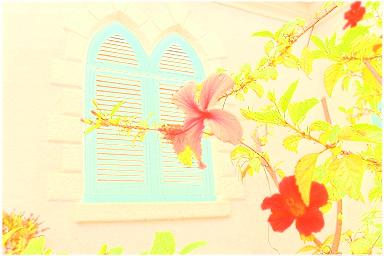}}
\centering{\scriptsize $\bm{R}, \alpha=10^{-4},\beta=10^{-4}$}
\end{subfigure}
\end{minipage}
\vspace{-1mm}
\caption{\linespread{1}\selectfont{Comparison of illumination $\bm{I}$ and reflectance $\bm{R}$ components by our STAR with different $\alpha,\beta$ on the ``kodim07'' image from the Kodak dataset, which is available at \url{http://r0k.us/graphics/kodak/kodim07.html}.}}
\label{F-AlphaBeta}
\end{figure*}

\noindent
\textbf{4. How to determine the parameters $\alpha$ and $\beta$?}
The scaling and relative size of these two parameters determine the trade-off of the regularization intensity between the structure and texture components.\ To determine their reasonable values, we run Retinex decomposition experiments on the ``kodim07'' image (described as ``a shuttered window partially masked by flowering bush'') by our STAR with $\alpha, \beta\in\{1, 10^{-1}, 10^{-2}, 10^{-3}, 10^{-4}\}$.\ To avoid extensive parameter tuning, We did not test our STAR with other values in $\alpha,\beta$.\ The original image can be available at \url{http://r0k.us/graphics/kodak/kodim07.html}.\ The decomposed illumination and reflectance components by our STAR with different $\alpha$ and $\beta$ values are shown in Figure~\ref{F-AlphaBeta}, respectively.\ Due to limited space, here we only show the components by our STAR with $\alpha\in\{10^{-1},10^{-2},10^{-3},10^{-4}\}$.\ Other comparisons by our STAR with $\alpha\in\{1\}$ and on other images are provided in the \textsl{Supplementary File}.\ We observe that our STAR with $\alpha=10^{-3}$ achieves better illumination performance than the other cases.\ By fixing the $\alpha=10^{-3}$, we observe that the reflectance reflects more details with smaller $\beta$.\ Similar results can be found on other images.\ Therefore, we set $\alpha=10^{-3}$ and $\beta=10^{-4}$ for our STAR.\

\noindent
\textbf{5. Is updating $\bm{S},\bm{T}$ necessary}?
We also study the effect of the updating iteration number $L$ on STAR.\ To do so, we simply set $L=1,2,4$ in STAR and evaluate its Retinex decomposition performance.\ From Figure~\ref{F-Validation-4}, one can see that the illumination becomes more structural while reflectance presents more details with more iterations.\

%-------------------------------------------------------------------------
\section{Other Applications}
\label{sec:appli}
In this section, we apply the proposed STAR model on two other image processing applications: low-light image enhancement (\S\ref{sec:enhance}) and color correction (\S\ref{sec:color}).\

\subsection{Low-light Image Enhancement}
\label{sec:enhance}
Capturing images in low-light environments suffers from unavoidable problems, such as low visibility~\cite{Liang_2018_CVPR} and heavy noise degradation~\cite{pgpd,mcwnnm,twsc}.\ Low-light image enhancement aims to alleviate this problem by improving the visibility and contrast of the observed images.\ To preserve the color information, the Retinex model based low-light image enhancement is often performed in the Value (V) channel of the Hue-Saturation-Value (HSV) domain.\

\noindent
\textbf{Comparison methods and datasets}.\ 
We compare the proposed STAR model with previous competing low-light image enhancement methods, including HE~\cite{he2004}, MSRCR~\cite{msrcr1997}, Contextual and  Variational Contrast (CVC)~\cite{cvc2011}, Naturalness Preserved Enhancement (NPE)~\cite{npe2013}, Layered Difference Representation (LDR)~\cite{ldr2013}, SIRE~\cite{srie2015}, Multi-scale Fusion (MF)~\cite{mf2016}, WVM~\cite{wvm2016}, Low-light IMage  Enhancement (LIME)~\cite{lime2017}, and JieP~\cite{jiep2017}.\ We evaluate these methods on 35 low-light images collected from~\cite{srie2015,wvm2016,lime2017,jiep2017}, and on the 200 low-light images provided in~\cite{npe2013}.

\noindent
\textbf{Objective metrics}.\
We qualitatively and quantitatively evaluate these methods on the subjective and objective quality metrics of enhanced images, respectively.\ The compared methods are evaluated on two commonly used metrics, one no-reference image quality assessment (IQA) metric Natural Image Quality Evaluator (NIQE)~\cite{niqe}, and one full-reference IAQ metric Visual Information Fidelity (VIF)~\cite{vif}.\ A lower NIQE value indicates better image quality, while a higher VIF value indicates better visual quality.\ The reason we employ VIF is that it is widely considered~\cite{wvm2016,jiep2017,lime2017} to capture visual quality better than the Peak Signal-to-Noise Ratio (PSNR) and the Structural Similarity Index (SSIM)~\cite{ssim}, which cannot be used in this task since no ``ground truth'' images are available.

\begin{figure*}
\vspace{-3mm}
\centering
\begin{minipage}[t]{0.233\textwidth}
\begin{subfigure}[t]{1\linewidth}
\raisebox{-0.15cm}{\includegraphics[width=1\textwidth]{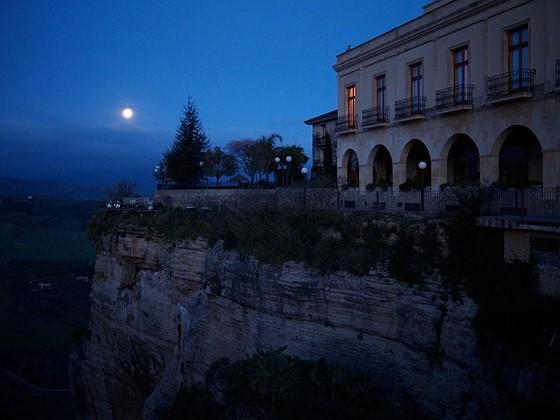}}
\end{subfigure}
\begin{subfigure}[t]{1\linewidth}
\raisebox{-0.15cm}{\includegraphics[width=1\textwidth]{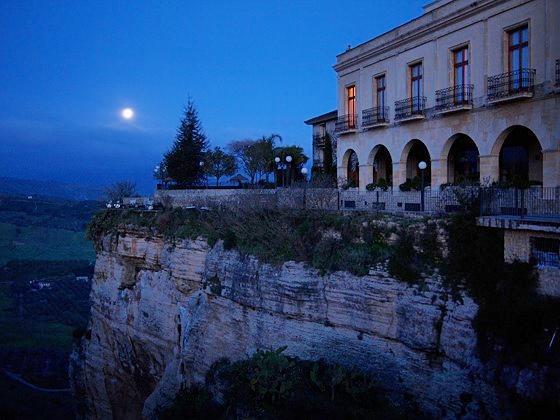}}
\end{subfigure}
\begin{subfigure}[t]{1\linewidth}
\raisebox{-0.15cm}{\includegraphics[width=1\textwidth]{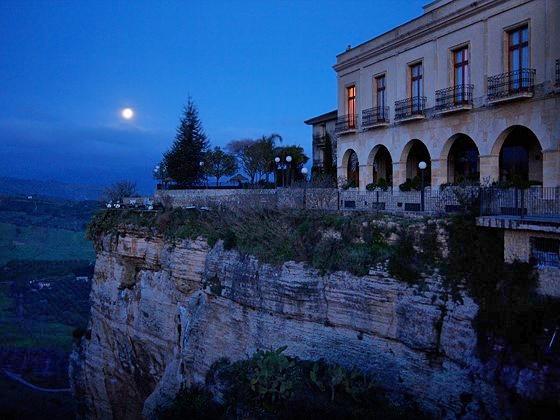}}
\end{subfigure}
\begin{subfigure}[t]{1\linewidth}
\raisebox{-0.15cm}{\includegraphics[width=1\textwidth]{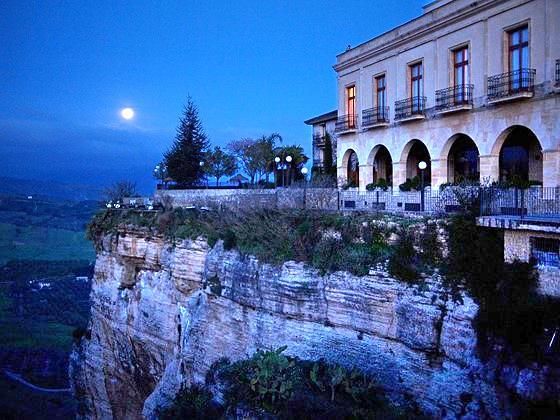}}
\end{subfigure}
\begin{subfigure}[t]{1\linewidth}
\raisebox{-0.15cm}{\includegraphics[width=1\textwidth]{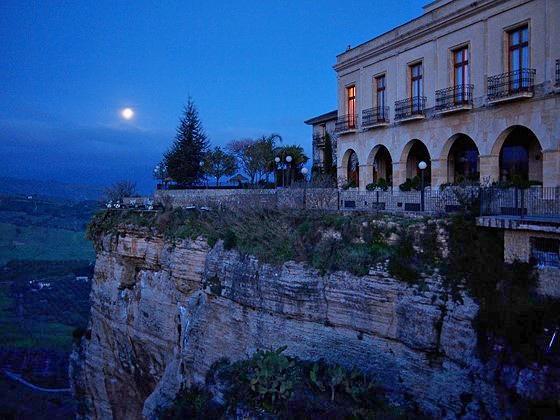}}
\end{subfigure}
\begin{subfigure}[t]{1\linewidth}
\raisebox{-0.15cm}{\includegraphics[width=1\textwidth]{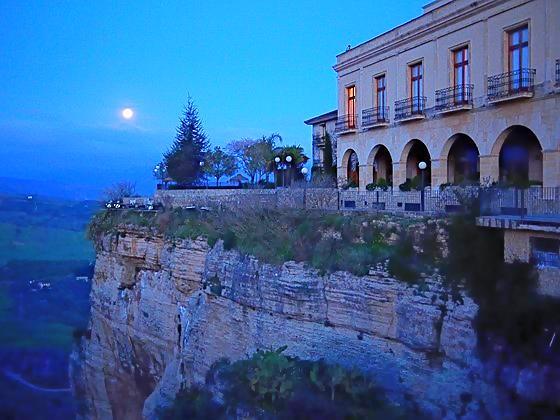}}
\end{subfigure}
\begin{subfigure}[t]{1\linewidth}
\raisebox{-0.15cm}{\includegraphics[width=1\textwidth]{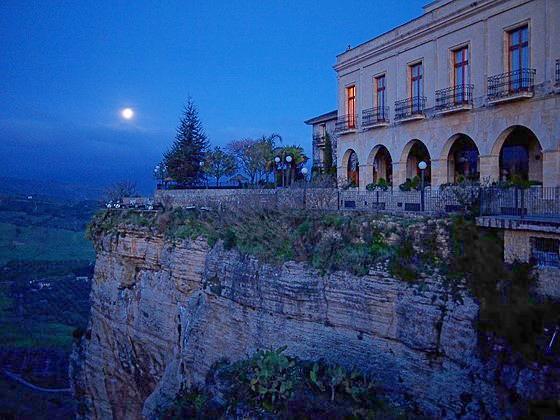}}
\end{subfigure}
\end{minipage}
\begin{minipage}[t]{0.156\textwidth}
\begin{subfigure}[t]{1\linewidth}
\raisebox{-0.15cm}{\includegraphics[width=1\textwidth]{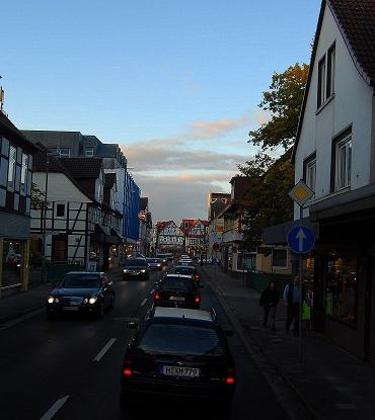}}
\end{subfigure}
\begin{subfigure}[t]{1\linewidth}
\raisebox{-0.15cm}{\includegraphics[width=1\textwidth]{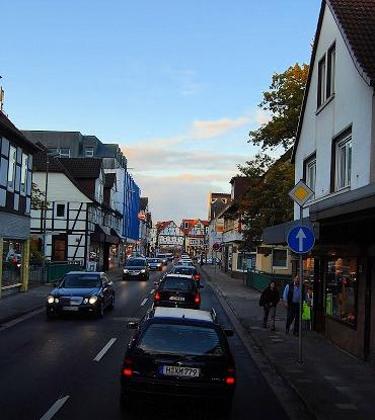}}
\end{subfigure}
\begin{subfigure}[t]{1\linewidth}
\raisebox{-0.15cm}{\includegraphics[width=1\textwidth]{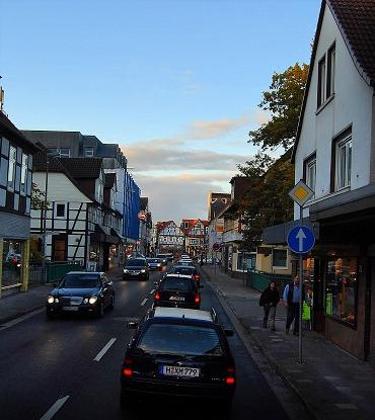}}
\end{subfigure}
\begin{subfigure}[t]{1\linewidth}
\raisebox{-0.15cm}{\includegraphics[width=1\textwidth]{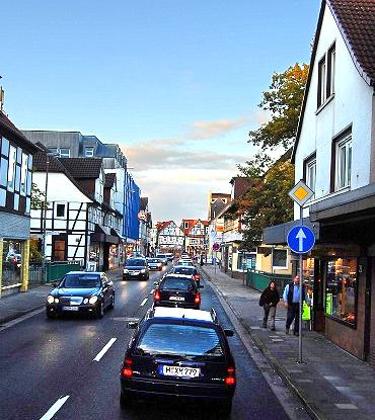}}
\end{subfigure}
\begin{subfigure}[t]{1\linewidth}
\raisebox{-0.15cm}{\includegraphics[width=1\textwidth]{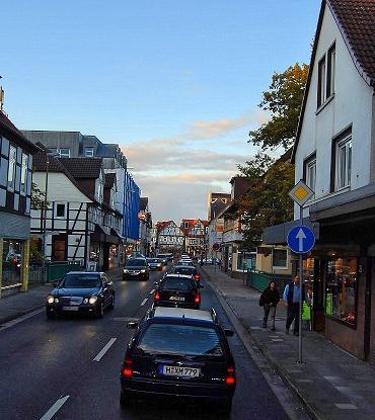}}
\end{subfigure}
\begin{subfigure}[t]{1\linewidth}
\raisebox{-0.15cm}{\includegraphics[width=1\textwidth]{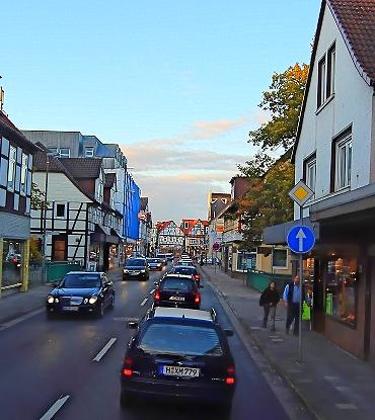}}
\end{subfigure}
\begin{subfigure}[t]{1\linewidth}
\raisebox{-0.15cm}{\includegraphics[width=1\textwidth]{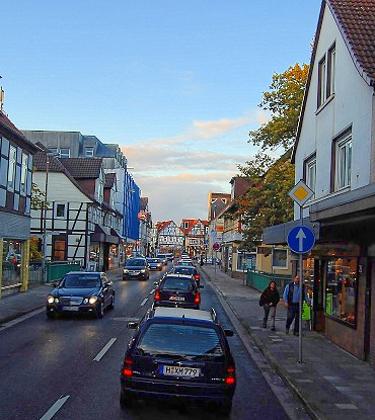}}
\end{subfigure}
\end{minipage}
\begin{minipage}[t]{0.1746\textwidth}
\begin{subfigure}[t]{1\linewidth}
\raisebox{-0.15cm}{\includegraphics[width=1\textwidth]{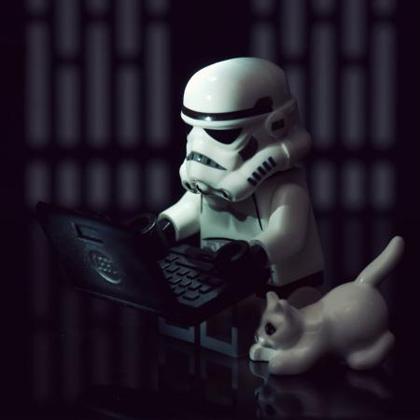}}
\end{subfigure}
\begin{subfigure}[t]{1\linewidth}
\raisebox{-0.15cm}{\includegraphics[width=1\textwidth]{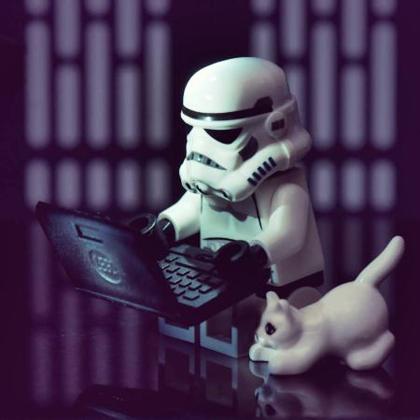}}
\end{subfigure}
\begin{subfigure}[t]{1\linewidth}
\raisebox{-0.15cm}{\includegraphics[width=1\textwidth]{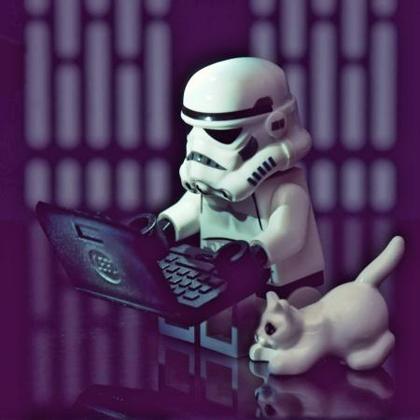}}
\end{subfigure}
\begin{subfigure}[t]{1\linewidth}
\raisebox{-0.15cm}{\includegraphics[width=1\textwidth]{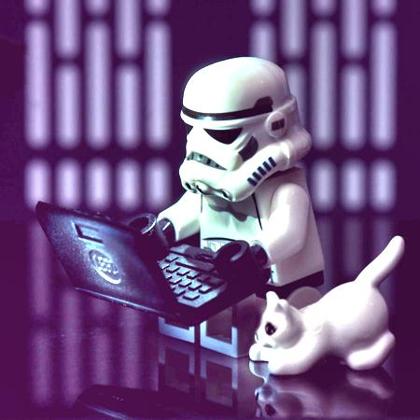}}
\end{subfigure}
\begin{subfigure}[t]{1\linewidth}
\raisebox{-0.15cm}{\includegraphics[width=1\textwidth]{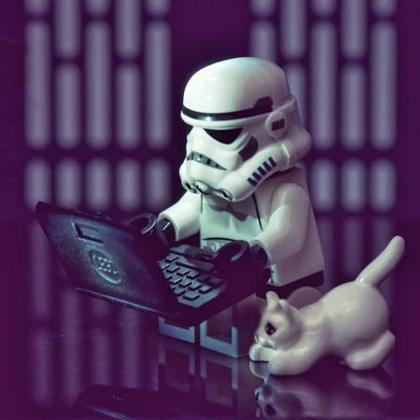}}
\end{subfigure}
\begin{subfigure}[t]{1\linewidth}
\raisebox{-0.15cm}{\includegraphics[width=1\textwidth]{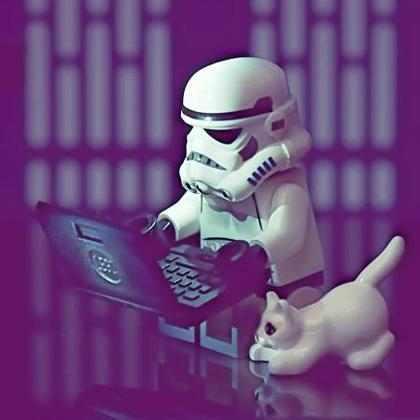}}
\end{subfigure}
\begin{subfigure}[t]{1\linewidth}
\raisebox{-0.15cm}{\includegraphics[width=1\textwidth]{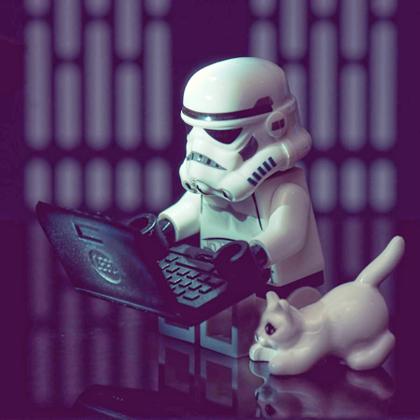}}
\end{subfigure}
\end{minipage}
\begin{minipage}[t]{0.1571\textwidth}
\begin{subfigure}[t]{1\linewidth}
\raisebox{-0.15cm}{\includegraphics[width=1\textwidth]{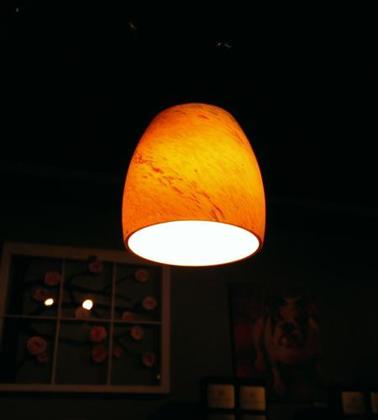}}
\end{subfigure}
\begin{subfigure}[t]{1\linewidth}
\raisebox{-0.15cm}{\includegraphics[width=1\textwidth]{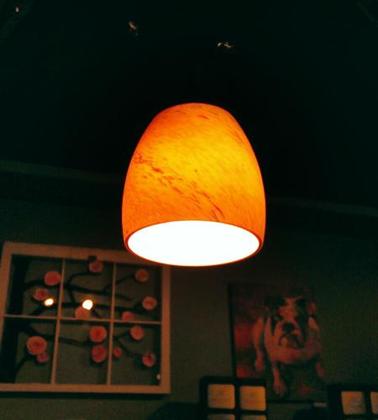}}
\end{subfigure}
\begin{subfigure}[t]{1\linewidth}
\raisebox{-0.15cm}{\includegraphics[width=1\textwidth]{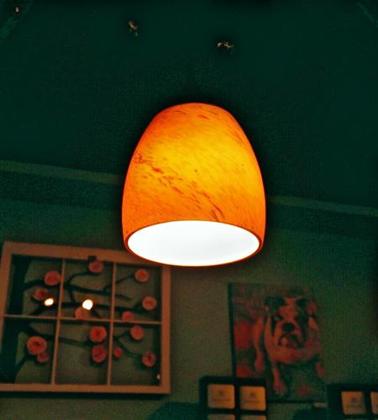}}
\end{subfigure}
\begin{subfigure}[t]{1\linewidth}
\raisebox{-0.15cm}{\includegraphics[width=1\textwidth]{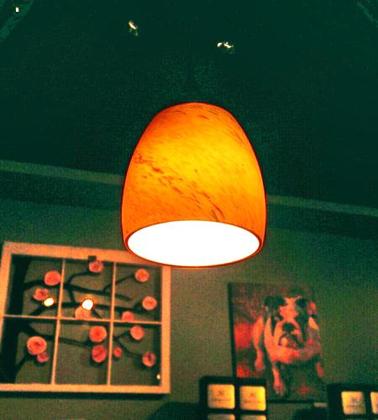}}
\end{subfigure}
\begin{subfigure}[t]{1\linewidth}
\raisebox{-0.15cm}{\includegraphics[width=1\textwidth]{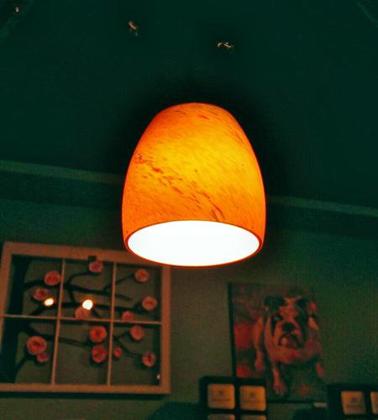}}
\end{subfigure}
\begin{subfigure}[t]{1\linewidth}
\raisebox{-0.15cm}{\includegraphics[width=1\textwidth]{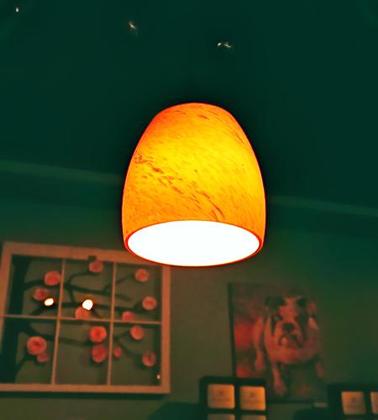}}
\end{subfigure}
\begin{subfigure}[t]{1\linewidth}
\raisebox{-0.15cm}{\includegraphics[width=1\textwidth]{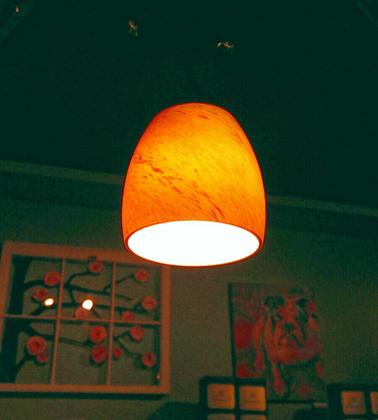}}
\end{subfigure}
\end{minipage}
\begin{minipage}[t]{0.23\textwidth}
\begin{subfigure}[t]{1\linewidth}
\raisebox{-0.15cm}{\includegraphics[width=1\textwidth]{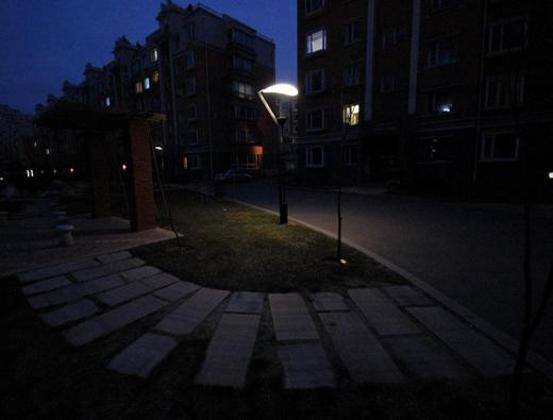}}
\end{subfigure}
\begin{subfigure}[t]{1\linewidth}
\raisebox{-0.15cm}{\includegraphics[width=1\textwidth]{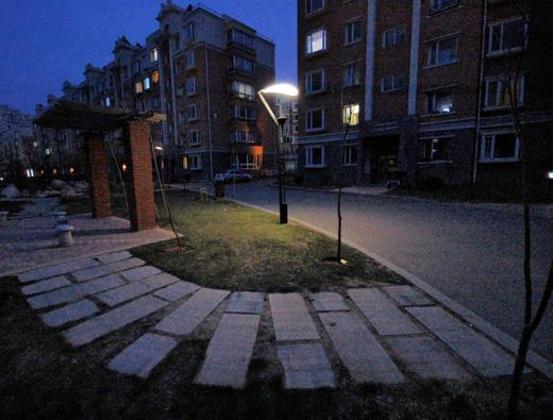}}
\end{subfigure}
\begin{subfigure}[t]{1\linewidth}
\raisebox{-0.15cm}{\includegraphics[width=1\textwidth]{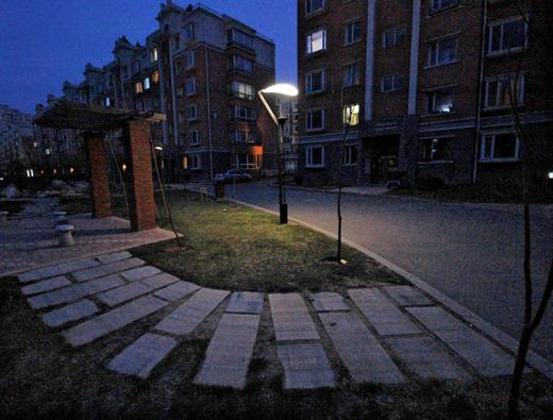}}
\end{subfigure}
\begin{subfigure}[t]{1\linewidth}
\raisebox{-0.15cm}{\includegraphics[width=1\textwidth]{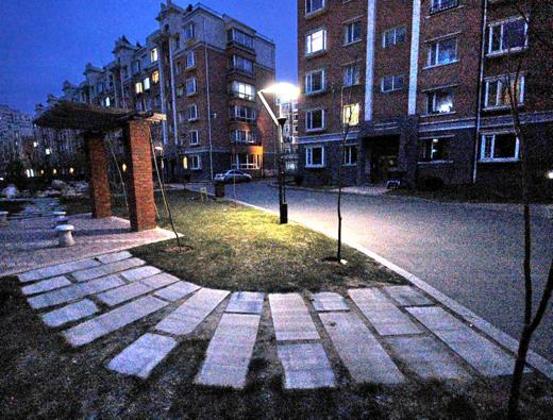}}
\end{subfigure}
\begin{subfigure}[t]{1\linewidth}
\raisebox{-0.15cm}{\includegraphics[width=1\textwidth]{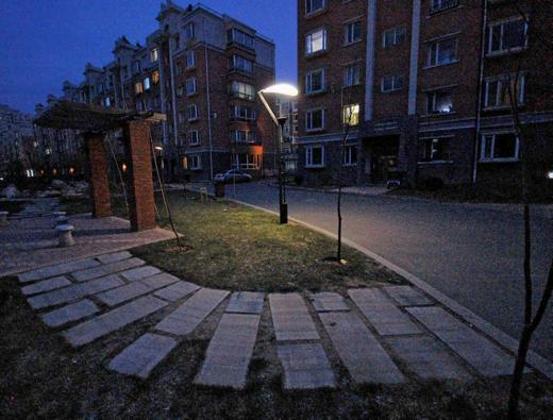}}
\end{subfigure}
\begin{subfigure}[t]{1\linewidth}
\raisebox{-0.15cm}{\includegraphics[width=1\textwidth]{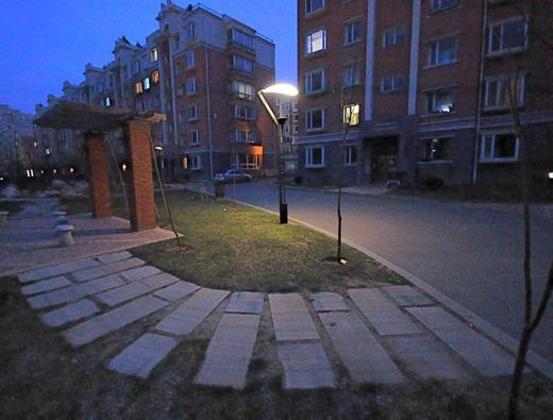}}
\end{subfigure}
\begin{subfigure}[t]{1\linewidth}
\raisebox{-0.15cm}{\includegraphics[width=1\textwidth]{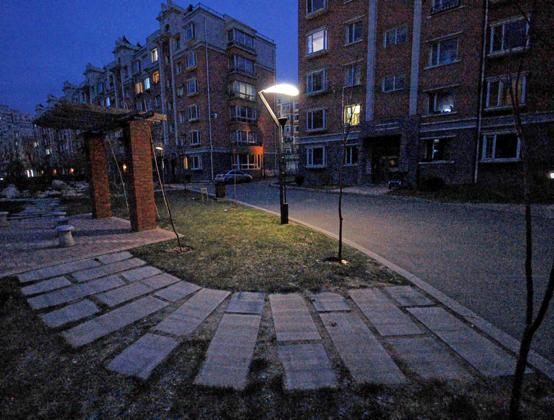}}
\end{subfigure}
\end{minipage}
\vspace{-1mm}
\caption{\linespread{1}\selectfont{Comparisons of enhanced images by different methods on 5 low-light images of the 35 images collected from~\cite{srie2015,wvm2016,lime2017,jiep2017}.\ The images from the first to the last rows are the original images, enhanced images by SIRE~\cite{srie2015}, WVM~\cite{wvm2016}, LIME~\cite{lime2017}, JieP~\cite{jiep2017}, RRM~\cite{li2018structure}, and our STAR, respectively.}}
\label{F-enhancement}
\end{figure*}

\noindent
\textbf{Results} are listed in Table~\ref{t-lowlight}.\ One can see that the proposed STAR achieves lower NIQE and higher VIF results than the other competing methods.\ This indicates that the images enhanced by our STAR present better visual quality than those of other methods.\ Besides, without the structure or texture weighting scheme, the proposed STAR model produces inferior performance on these two objective metrics.\ This demonstrates the effectiveness of the proposed structure and texture aware components for low-light image enhancement.\ In Figure~\ref{F-enhancement}, we compare the visual quality of state-of-the-art methods~\cite{srie2015,wvm2016,lime2017,jiep2017,li2018structure}.\ As can be seen, on several representative images, our STAR achieves visually clear content while enhancing the illumination naturally, in agreement with our objective results.\ Besides, from the $5$-th column of Figure~\ref{F-enhancement}, one can observe that the proposed STAR achieves comparable performance with the competing methods~\cite{wvm2016,lime2017,jiep2017} on noise suppression~\cite{guide,nlh2020,ren2019simultaneous}.

\begin{table}
\begin{center}
\footnotesize
\begin{tabular}{r|c|c|c|c}
\Xhline{1pt}
\rowcolor[rgb]{ .851,  .851,  .851}
Dataset 
& \multicolumn{2}{c|}
{35 Images}
& \multicolumn{2}{c}
{200 Images} 
\\
%\hline
\rowcolor[rgb]{ .851,  .851,  .851}
Metric 
& NIQE $\downarrow$ & VIF $\uparrow$ & NIQE $\downarrow$ & VIF $\uparrow$ 
\\
\hline\hline
\textbf{Input} & 3.74 & 1.00 & 3.45 & 1.00 
\\
\textbf{HE}~\cite{he2004}& 3.24 & 1.34 & 3.28 & 1.19 
\\
\textbf{MSRCR}~\cite{msrcr1997} & 2.98 & 1.84 & 3.21 & 1.11 
\\
\textbf{CVC}~\cite{cvc2011} & 3.03 & 2.04 & 3.01 & 1.63
\\
\textbf{NPE}~\cite{npe2013}& 3.10 & 2.48  & 3.12 & 1.62
\\
\textbf{LDR}~\cite{ldr2013} & 3.12 & 2.36 & 2.96 & 1.66
\\
\textbf{SIRE}~\cite{srie2015}& 3.06 & 2.09  & 2.98 & 1.57 
\\
\textbf{MF}~\cite{mf2016} & 3.19 & 2.23  & 3.26 & 1.71 
\\
\textbf{WVM}~\cite{wvm2016} & 2.98 & 2.22 & 2.99 & 1.68
\\
\textbf{LIME}~\cite{lime2017} & 3.24 & 2.76 & 3.32 & 1.84
\\
\textbf{JieP}\ \ ~\cite{jiep2017} & 3.06 & 2.67 & 3.18 & 1.82
\\
\textbf{BIMEF}~\cite{bimef} & 3.14 & 2.54 & 3.02 & 1.79
\\
\textbf{RRM}~\cite{li2018structure} & 3.08 & 2.69 & 2.97 & 1.86
\\
\hline
\hline
\multicolumn{1}{l|}{\textbf{STAR w/o $\bm{S}$}} 
& 3.18 & 2.64 & 3.22 & 1.77
\\
\multicolumn{1}{l|}{\textbf{STAR w/o $\bm{T}$}} 
& 3.09 & 2.78 & 3.01 & 1.82
\\
\hline
\multicolumn{1}{l|}{\textbf{STAR}}
& \textbf{2.93} & \textbf{2.96} & \textbf{2.86} & \textbf{1.92}
\\
\hline
\end{tabular}
\end{center}
\vspace{-2mm}
\caption{Average NIQE~\cite{niqe} and VIF~\cite{vif} results of different methods on 35 low-light images collected from~\cite{srie2015,wvm2016,lime2017,jiep2017} and 200 low-light images provided in~\cite{npe2013}.}
\vspace{-2mm}
\label{t-lowlight}
\end{table}

\begin{figure*}
\centering
\begin{minipage}[t]{0.245\textwidth}
\begin{subfigure}[t]{1\linewidth}
\raisebox{-0.15cm}{\includegraphics[width=1\textwidth]{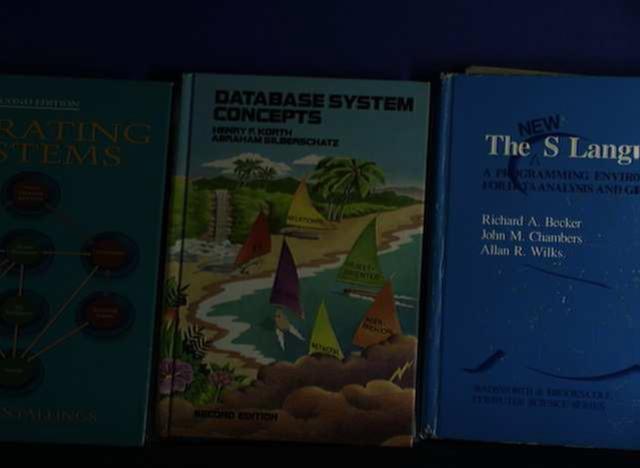}}
\end{subfigure}
\begin{subfigure}[b]{1\linewidth}
\raisebox{-0.15cm}{\includegraphics[width=1\textwidth]{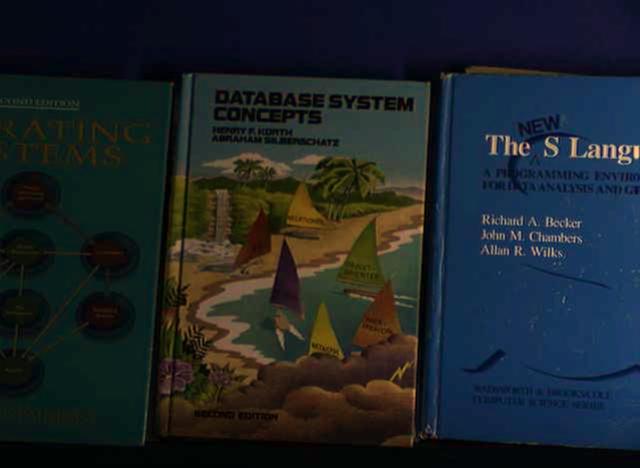}}
\end{subfigure}
\begin{subfigure}[b]{1\linewidth}
\raisebox{-0.15cm}{\includegraphics[width=1\textwidth]{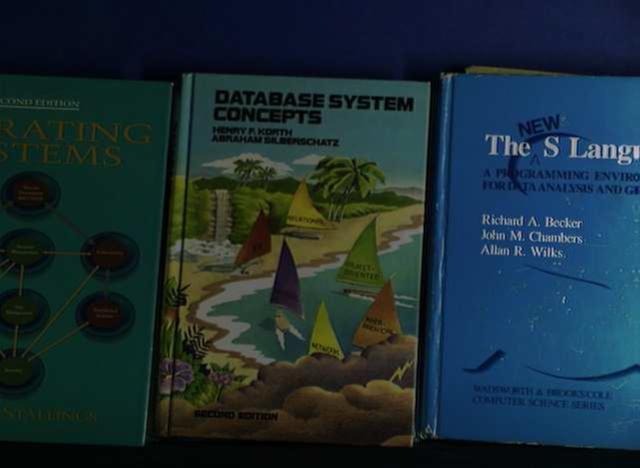}}
\end{subfigure}
\begin{subfigure}[b]{1\linewidth}
\raisebox{-0.15cm}{\includegraphics[width=1\textwidth]{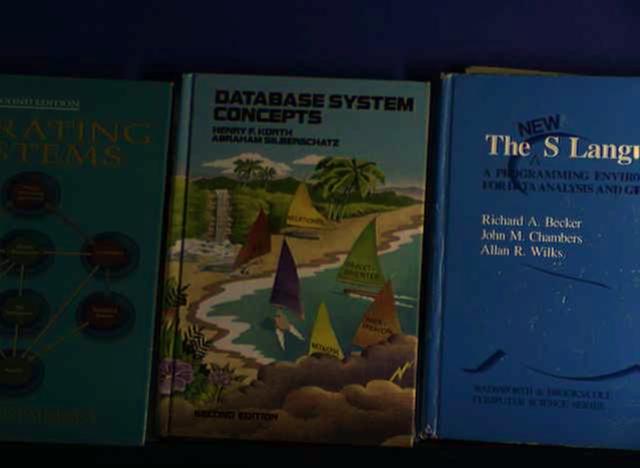}}
\end{subfigure}
\begin{subfigure}[b]{1\linewidth}
\raisebox{-0.15cm}{\includegraphics[width=1\textwidth]{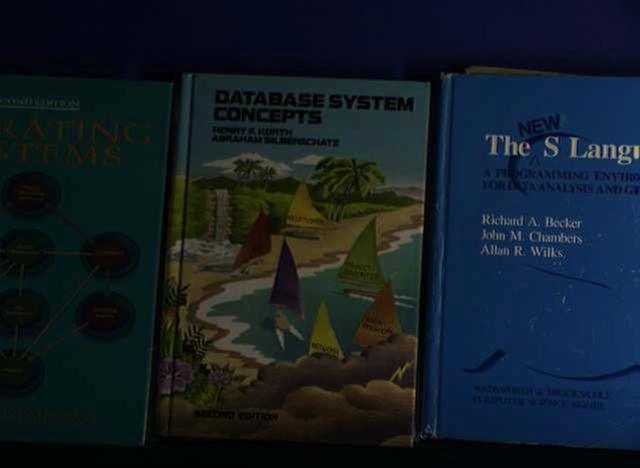}}
\end{subfigure}j
\begin{subfigure}[b]{1\linewidth}
\raisebox{-0.15cm}{\includegraphics[width=1\textwidth]{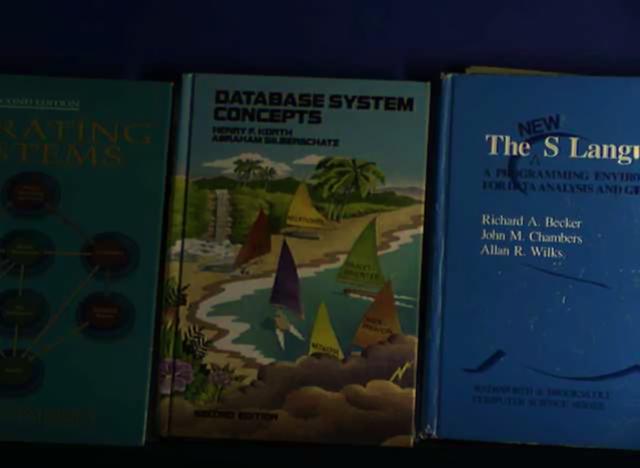}}
\end{subfigure}
\end{minipage}
\begin{minipage}[t]{0.245\textwidth}
\begin{subfigure}[t]{1\linewidth}
\raisebox{-0.15cm}{\includegraphics[width=1\textwidth]{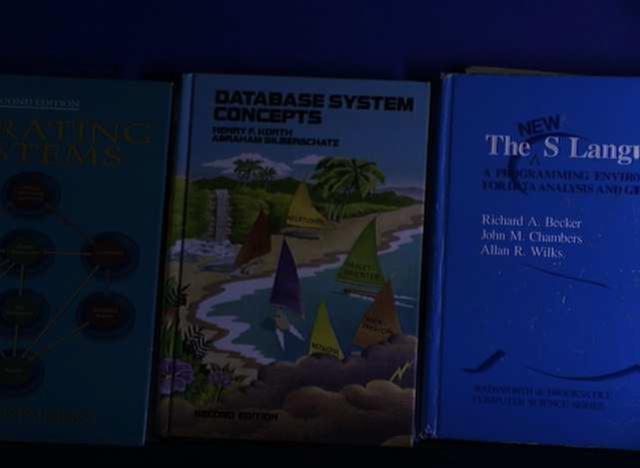}}
\end{subfigure}
\begin{subfigure}[b]{1\linewidth}
\raisebox{-0.15cm}{\includegraphics[width=1\textwidth]{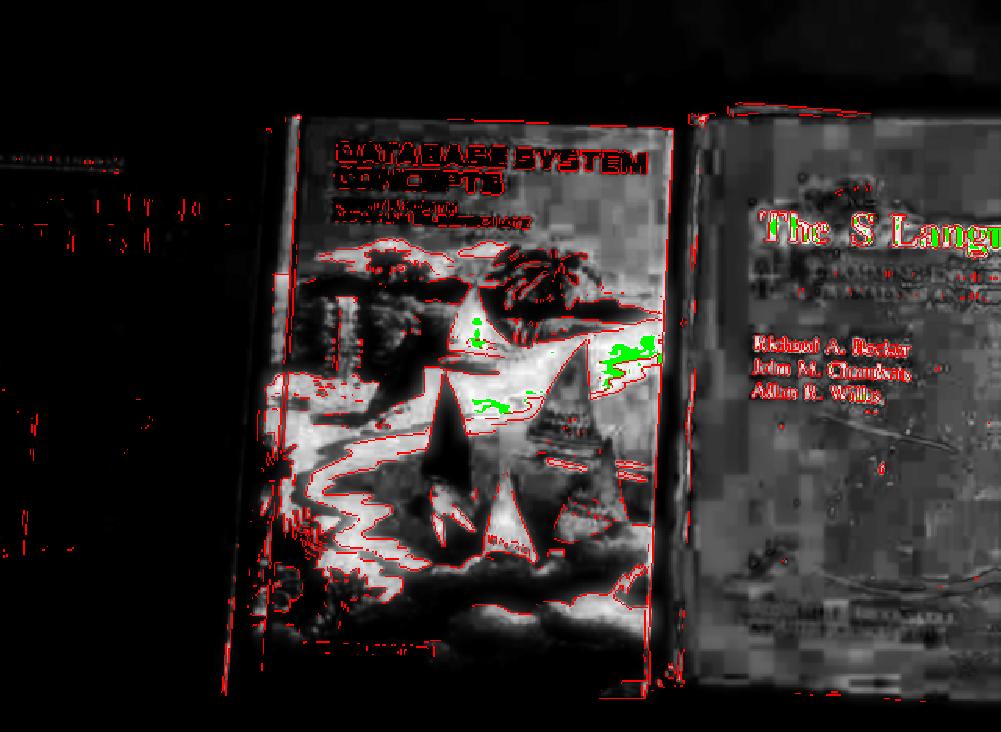}}
\end{subfigure}
\begin{subfigure}[b]{1\linewidth}
\raisebox{-0.15cm}{\includegraphics[width=1\textwidth]{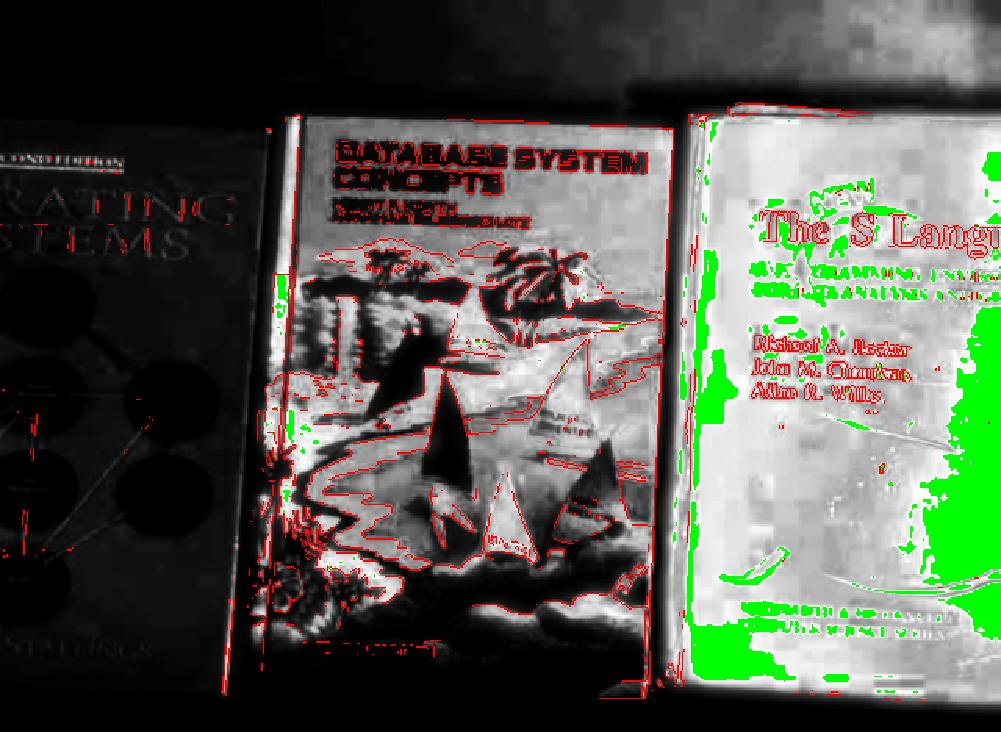}}
\end{subfigure}
\begin{subfigure}[b]{1\linewidth}
\raisebox{-0.15cm}{\includegraphics[width=1\textwidth]{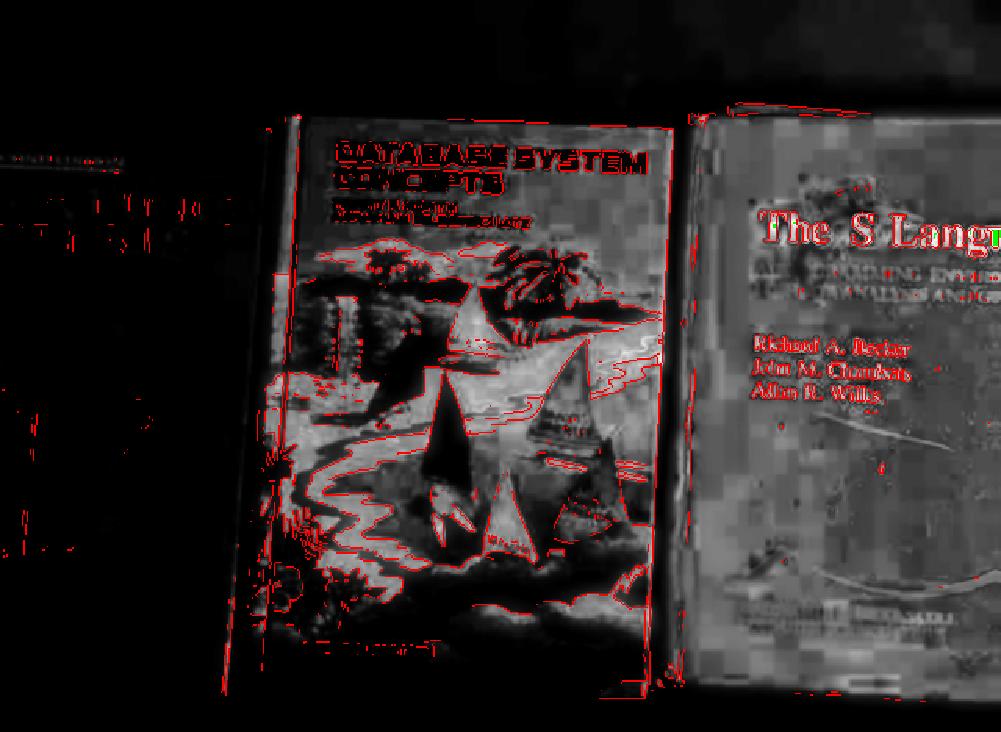}}
\end{subfigure}
\begin{subfigure}[b]{1\linewidth}
\raisebox{-0.15cm}{\includegraphics[width=1\textwidth]{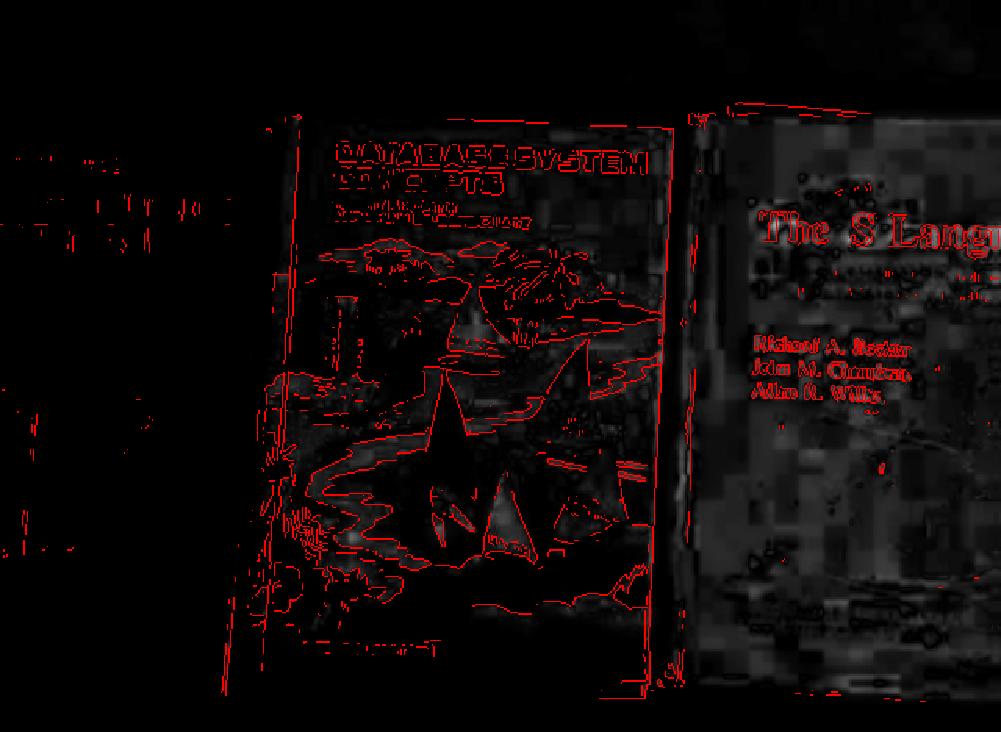}}
\end{subfigure}
\begin{subfigure}[b]{1\linewidth}
\raisebox{-0.15cm}{\includegraphics[width=1\textwidth]{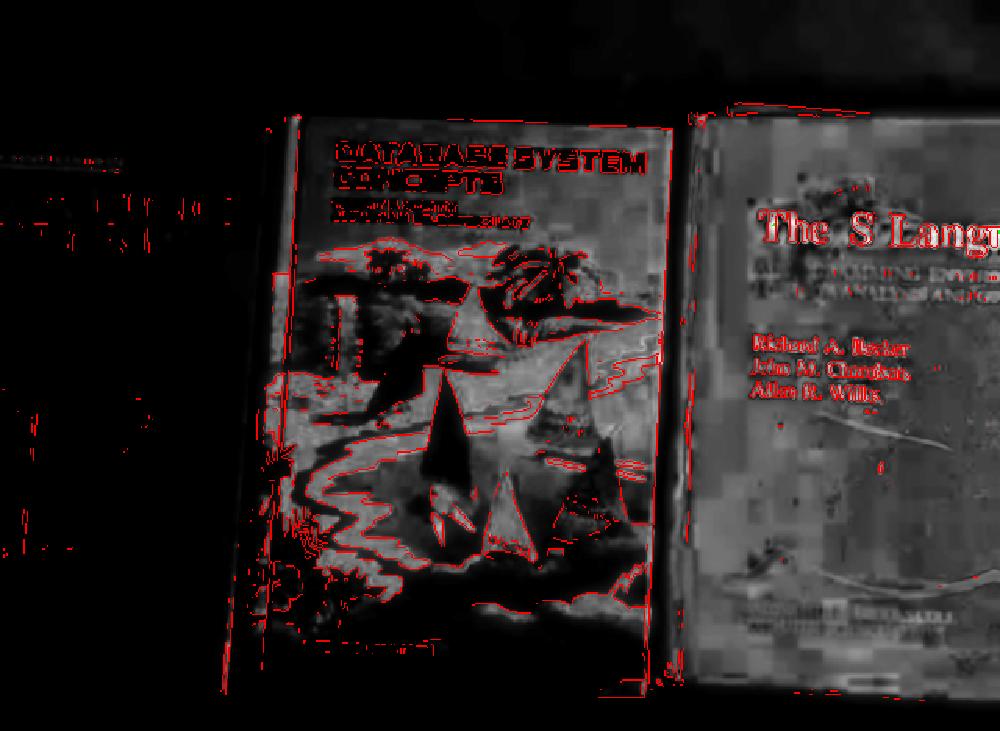}}
\end{subfigure}
\end{minipage}
\begin{minipage}[t]{0.245\textwidth}
\begin{subfigure}[t]{1\linewidth}
\raisebox{-0.15cm}{\includegraphics[width=1\textwidth]{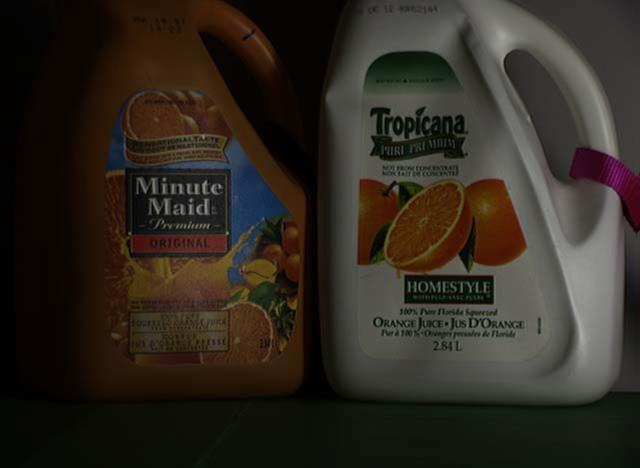}}
\end{subfigure}
\begin{subfigure}[b]{1\linewidth}
\raisebox{-0.15cm}{\includegraphics[width=1\textwidth]{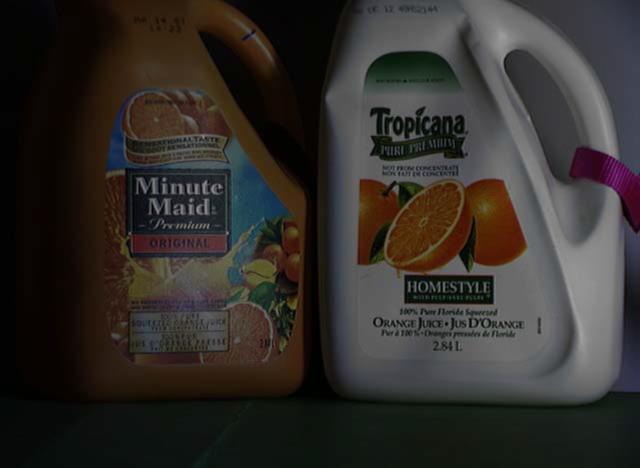}}
\end{subfigure}
\begin{subfigure}[b]{1\linewidth}
\raisebox{-0.15cm}{\includegraphics[width=1\textwidth]{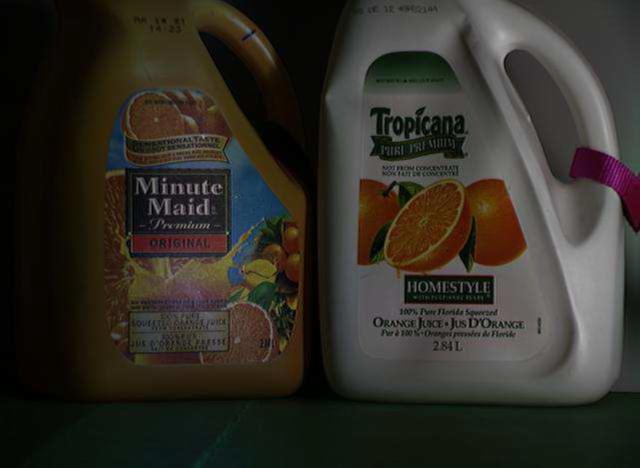}}
\end{subfigure}
\begin{subfigure}[b]{1\linewidth}
\raisebox{-0.15cm}{\includegraphics[width=1\textwidth]{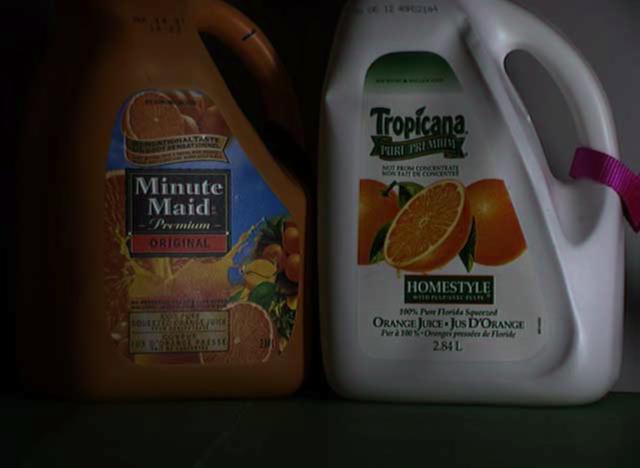}}
\end{subfigure}
\begin{subfigure}[b]{1\linewidth}
\raisebox{-0.15cm}{\includegraphics[width=1\textwidth]{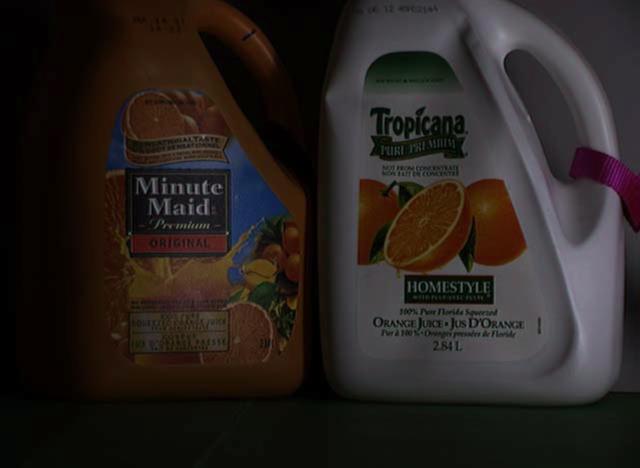}}
\end{subfigure}j
\begin{subfigure}[b]{1\linewidth}
\raisebox{-0.15cm}{\includegraphics[width=1\textwidth]{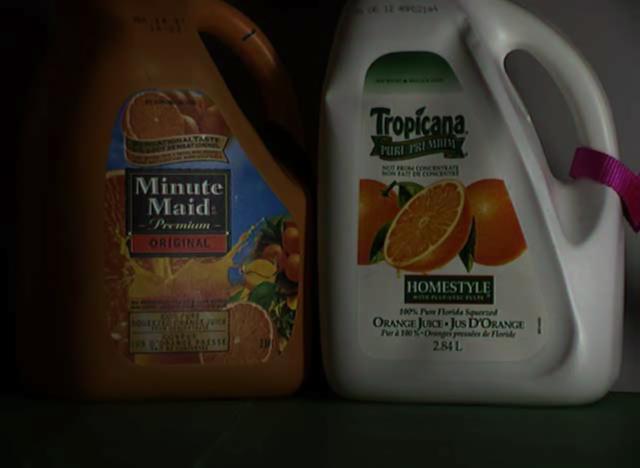}}
\end{subfigure}
\end{minipage}
\begin{minipage}[t]{0.245\textwidth}
\begin{subfigure}[t]{1\linewidth}
\raisebox{-0.15cm}{\includegraphics[width=1\textwidth]{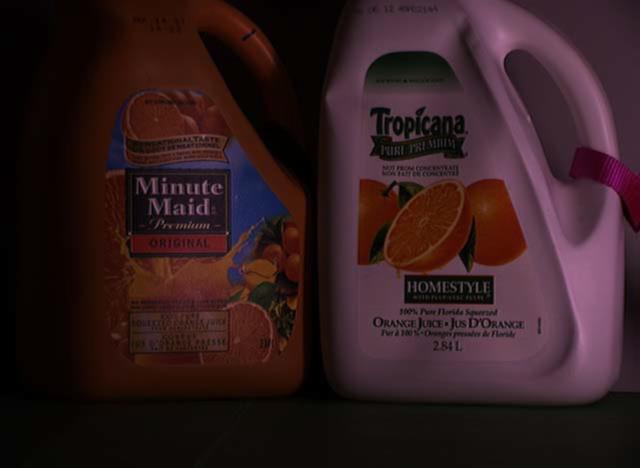}}
\end{subfigure}
\begin{subfigure}[b]{1\linewidth}
\raisebox{-0.15cm}{\includegraphics[width=1\textwidth]{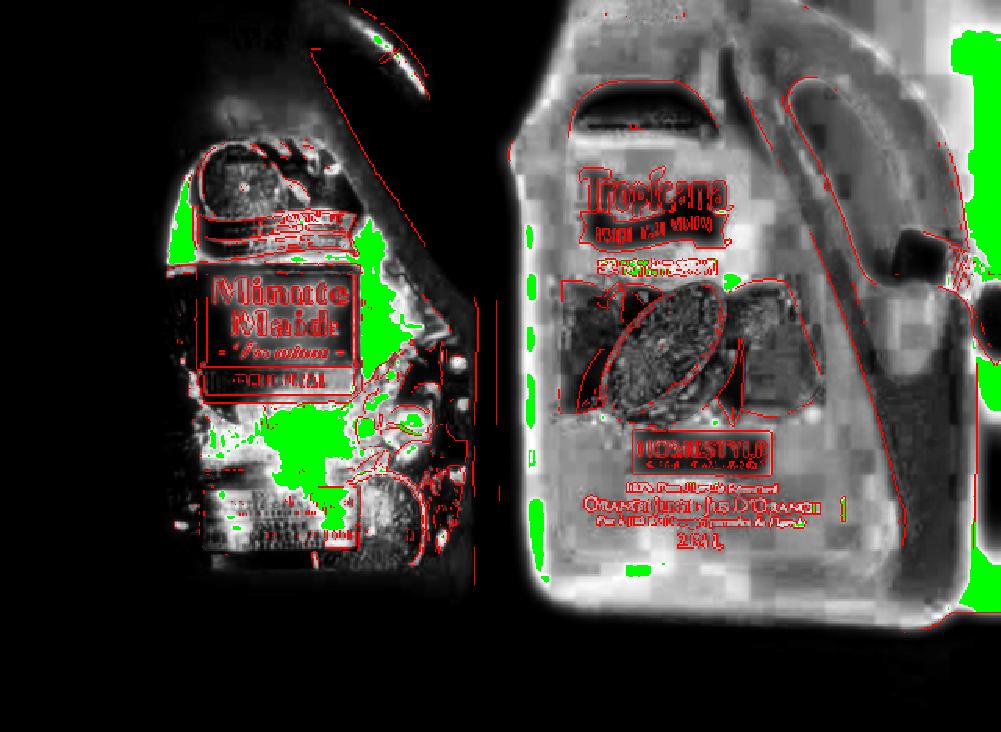}}
\end{subfigure}
\begin{subfigure}[b]{1\linewidth}
\raisebox{-0.15cm}{\includegraphics[width=1\textwidth]{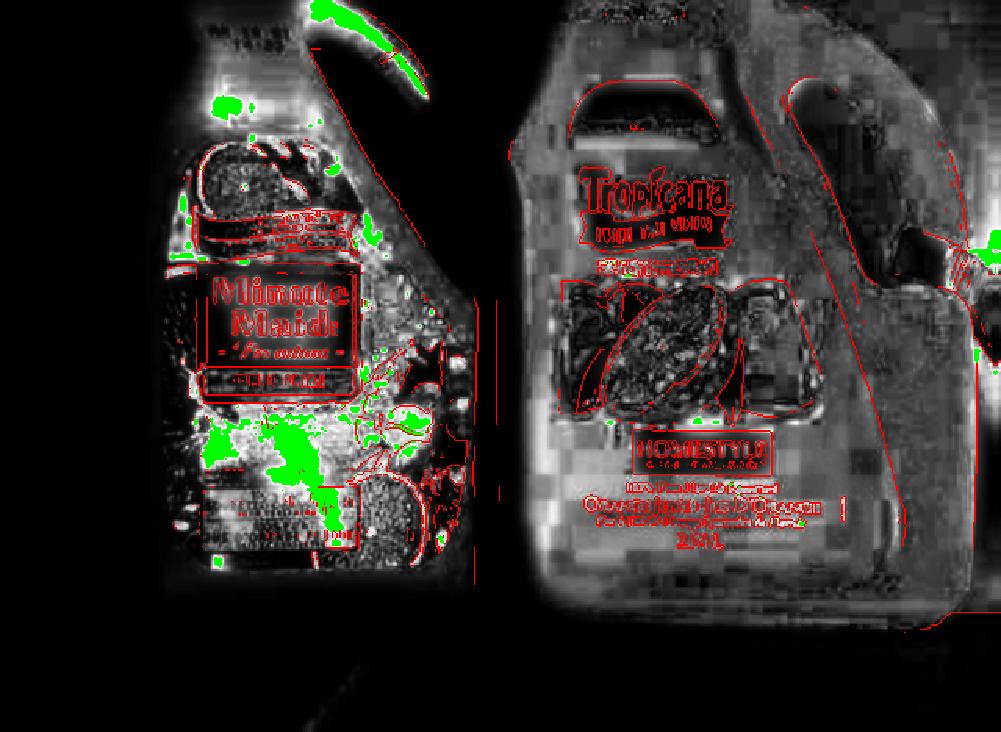}}
\end{subfigure}
\begin{subfigure}[b]{1\linewidth}
\raisebox{-0.15cm}{\includegraphics[width=1\textwidth]{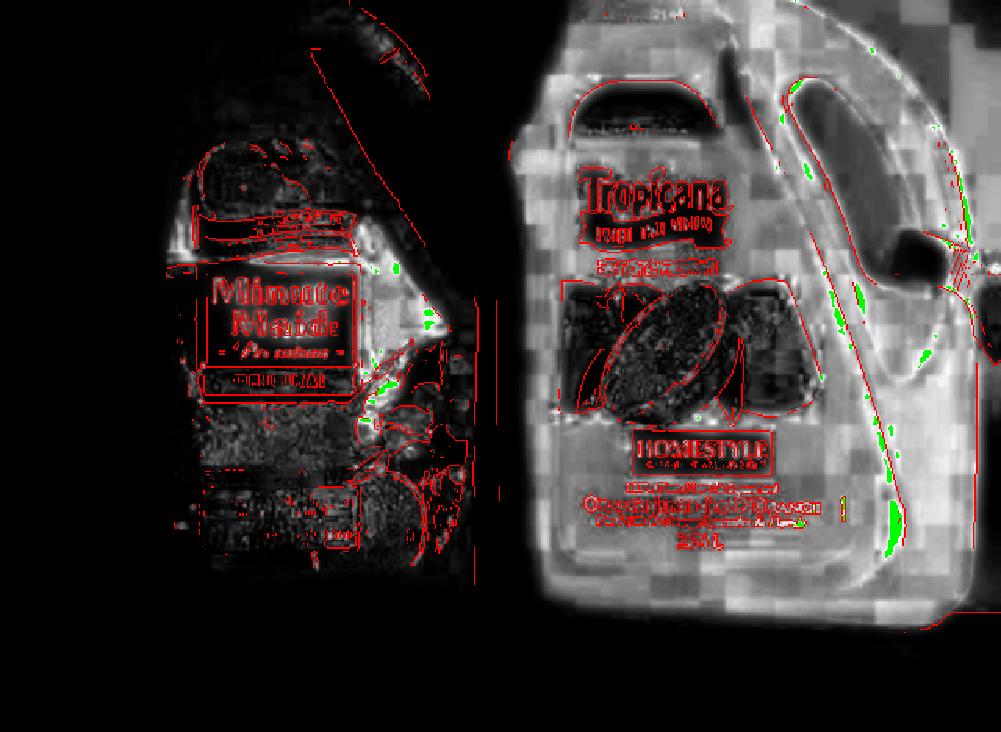}}
\end{subfigure}
\begin{subfigure}[b]{1\linewidth}
\raisebox{-0.15cm}{\includegraphics[width=1\textwidth]{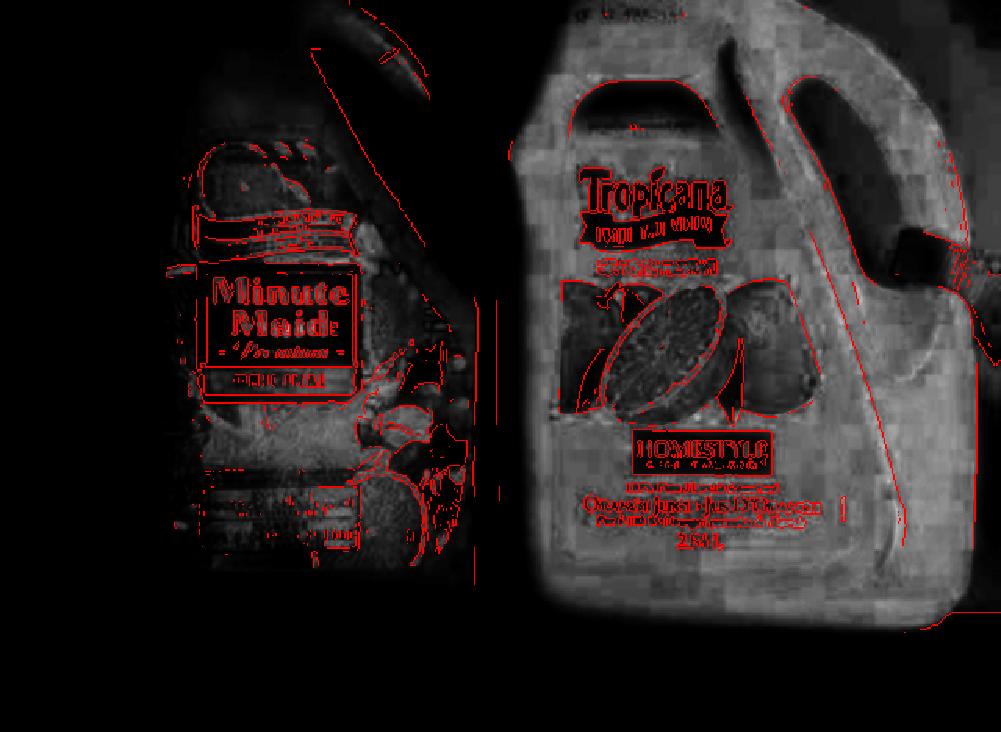}}
\end{subfigure}
\begin{subfigure}[b]{1\linewidth}
\raisebox{-0.15cm}{\includegraphics[width=1\textwidth]{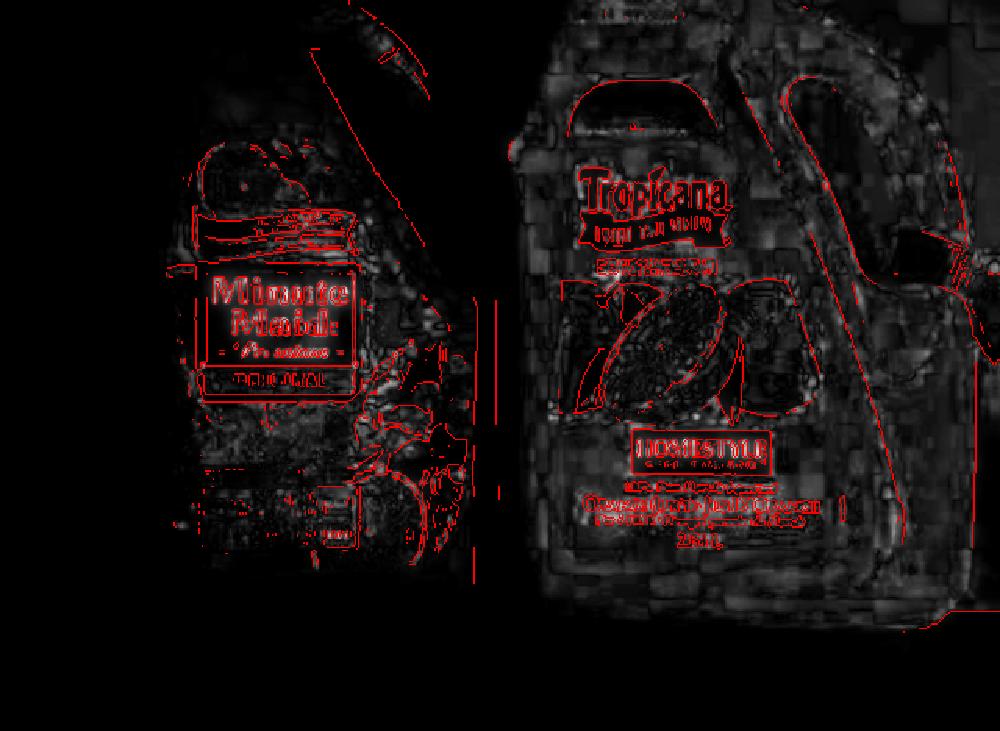}}
\end{subfigure}
\end{minipage}
\caption{\linespread{1}\selectfont{Visual comparisons of the ground truths ($1$-st and $3$-rd columns, $1$-st row) and SIRE~\cite{srie2015} ($2$-nd row), WVM~\cite{wvm2016} ($3$-rd row), JieP~\cite{jiep2017} ($4$-th row), LSRS~\cite{lsrs} ($5$-th row), and our STAR ($6$-th row) on color correction, as well as the corresponding input image ($2$-nd and $4$-th columns, $1$-st row) and the S-CIELAB errors ($2$-nd and $4$-th columns).}}
\label{F-ColorCorrection}
\end{figure*}

%-------------------------------------------------------------------------
\subsection{Color Correction}
\label{sec:color}
In Retinex theory~\cite{land1971lightness,land1977retinex}, if the estimation is performed in each channel of the RGB-color space, the estimated reflectance contains the original color information of the observed scene.\ Therefore, the Retinex model can be applied to color correction tasks.\ To demonstrate the estimation accuracy of the illumination and reflectance components, we evaluate the color correction performance of the proposed STAR model and the competing methods~\cite{srie2015,wvm2016,jiep2017}.

We first compare the performance of the proposed STAR with several leading Retinex methods: SIRE~\cite{srie2015}, WVM~\cite{wvm2016}, JieP~\cite{jiep2017}, and LSRS~\cite{lsrs}.\ The original images and color corrected images are downloaded from the \href{http://colorconstancy.com/?page_id=21}{Color Constancy Website}.\ In Figure~\ref{F-ColorCorrection}, we provide some visual results of color correction using different methods.\ One can see that, all these methods achieve satisfactory qualitative performance (from $2$-nd to $6$-th rows, $1$-st and $3$-rd columns of Figure~\ref{F-ColorCorrection}), when compared with the original images ($1$-st row, $2$-nd and $4$-th columns of Figure~\ref{F-ColorCorrection}) and ground truth images ($1$-st row, $1$-st and $3$-rd columns of Figure~\ref{F-ColorCorrection}).\ To verify the accuracy of color correction using these methods, we employ the S-CIELAB color metric~\cite{scielab} to measure the color errors on spatial processing.\ The S-CIELAB errors between the ground truth and corrected images of different methods are shown from the $2$-nd to $6$-th rows, $2$-nd and $4$-th columns of Figure~\ref{F-ColorCorrection}, respectively.\ As can be seen, the spatial locations of the errors, i.e., the green areas, of the STAR corrected images are much smaller than other methods.\ This indicates that the results of STAR are closer to the ground truth images ($1$-st row, $1$-st and $3$-rd columns of Figure~\ref{F-ColorCorrection}) than other methods.

Furthermore, we perform a quantitative comparison of the proposed STAR with several leading color constancy methods~\cite{bianco2015color,yang2015efficient,jiep2017} on the Color-Checker dataset~\cite{gehler2008bayesian}.\ This dataset contains totally 568 images of indoor and outdoor scenes taken with two high quality cameras (Canon 5D and Canon 1D).\ Each image contains a MacBeth color-checker for accuracy reference.\ The average illumination across each channel is computed in the RGB-color space separately, as the estimated illumination for that channel.\ The results in terms of Mean Angular Error (MAE, lower is better) between the corrected image and the ground truth image are listed in Table~\ref{T-ColorChecker}.\ As can be seen, the proposed STAR method achieves lower MAE results than the competing methods on the color constancy problem.

%----------------------------
\section{Conclusion}
\label{sec:conclusion}

In this paper, we proposed a Structure and Texture Aware Retinex (STAR) model for illumination and reflectance decomposition.\ We first introduced an Exponentialized Mean Local Variance (EMLV) filter to extract the structure and texture maps from the observed image.\ The extracted maps were employed to regularize the illumination and reflectance components.\ In addition, we proposed to alternatively update the structure/texture maps, and estimate the illumination/reflectance for better Retinex decomposition performance.\ The proposed STAR model is efficiently solved by a standard alternative optimization algorithm.\ Comprehensive experiments on Retinex decomposition, low-light image enhancement, and color correction demonstrated that the proposed STAR model achieves better quantitative and qualitative performance than representative Retinex decomposition methods.

Current Retinex decomposition community also has its problem.\ To the best of our knowledge, there is no reasonable ground truths of decomposed illumination and reflectance in Retinex decomposition.\ Most of the previous Retinex methods~\cite{srie2015,li2018structure,rdgan} compared the visual quality of decomposed illumination and reflectance components by subjective evaluations.\ In our opinion, the reasons are possibly two-fold: 1) it is hard to synthesis the illumination and reflectance components simultaneously to produce a meaningful natural image (just like the ``chicken or the egg'' problem~\cite{Borji2019,Zhao2018});\ 2) designing meaningful quantitative metrics (besides of PSNR and SSIM~\cite{ssim}) for the ground truth components (if have) are also difficult~\cite{Wang2019}, since both the illumination and reflectance components should be considered to evaluate the performance of Retinex methods.\ To perform quantitative comparisons, these methods performed experiments on low-light image enhancement and color correction, which largely depends on the quality of decomposed components, to demonstrate the advantages of the developed Retinex decomposition methods.\ How to directly benchmark existing Retinex methods on synthetic datasets with ground truths is really a challenging problem, and we will explore this direction as our future work.
\begin{table}
\begin{center}
\footnotesize
\begin{tabular}{c||cccccc}
\Xhline{1pt}
\rowcolor[rgb]{ .851,  .851,  .851}
Method 
& 
Gray-Edge~\cite{van2007edge}
& 
Shades-Gray~\cite{finlayson2004shades}
&
Bayesian~\cite{gehler2008bayesian}
\\
MAE$\downarrow$ 
& 5.13 & 4.93 & 4.82
\\
\hline
\rowcolor[rgb]{ .851,  .851,  .851}
Method
&
CNNs~\cite{bianco2015color}
&
Gray-World~\cite{buchsbaum1980spatial}
&
Gray-Pixel~\cite{yang2015efficient}
\\
MAE$\downarrow$ 
& 4.73 & 4.66 & 4.60 
\\
\hline
\rowcolor[rgb]{ .851,  .851,  .851}
Method
&
LSRS~\cite{lsrs}
&
JieP~\cite{jiep2017}
&
STAR
\\
MAE$\downarrow$ & 4.39 & 4.32 & \textbf{4.11}
\\
\hline
\end{tabular}
\end{center}
\vspace{-2mm}
\caption{Comparisons of different methods on Mean Angular Errors (MAE) on the Color-Checker dataset~\cite{gehler2008bayesian}.}
\label{T-ColorChecker}
\end{table}

{
%\balance
\bibliographystyle{IEEE} %unsrt
\bibliography{star.bib}
}

\vspace{-15mm}
\begin{IEEEbiography}[{\includegraphics[width=1in,height=1in,clip,keepaspectratio]{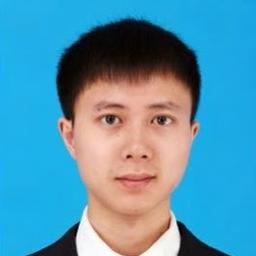}}]{Jun Xu} is an Assistant Professor in College of Computer Science, Nankai University, Tianjin, China.\
He received the B.Sc. degree in Pure Mathematics and the M. Sc. Degree in Information and Probability both from the School of Mathematics Science, Nankai University, China, in 2011 and 2014, respectively.\ He received the Ph.D. degree in the Department of Computing, The Hong Kong Polytechnic University in 2018.\ He worked as a Research Scientist in Inception Institute of Artificial Intelligence (IIAI), Abu Dhabi, UAE.\ More information can be found on his homepage \url{https://csjunxu.github.io/}. 
\end{IEEEbiography}

\vspace{-15mm}
\begin{IEEEbiography}[{\includegraphics[width=1in,height=1in,clip,keepaspectratio]{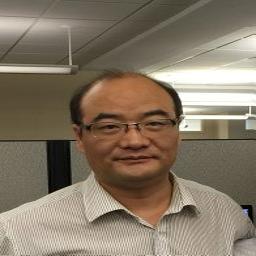}}]{Yingkun Hou} is an Associate Professor in School of Information Science and Technology, Taishan University, Taian, China.
\end{IEEEbiography}

\vspace{-15mm}
\begin{IEEEbiography}[{\includegraphics[width=1in,height=1in,clip,keepaspectratio]{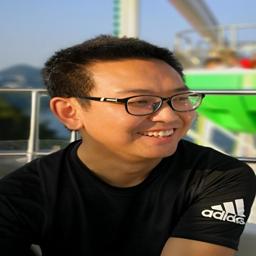}}]{Dongwei Ren} is an Assistant Professor in the College of Intelligence and Computing, Tianjin University, Tianjin, China.\ He received the Bachelor degree and the Master degree from Harbin Institute of Technology in 2011 and 2013. He received two Ph.D. degrees from Harbin Institute of Technology and The Hong Kong Polytechnic University in 2017 and 2018.
\end{IEEEbiography}

\vspace{-15mm}
\begin{IEEEbiography}[{\includegraphics[width=1in,height=1in,clip,keepaspectratio]{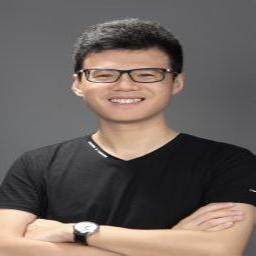}}]{Li Liu} is with Inception Institute of Artificial Intelligence (IIAI), Abu Dhabi, UAE.\
\end{IEEEbiography}

\vspace{-15mm}
\begin{IEEEbiography}[{\includegraphics[width=1in,height=1in,clip,keepaspectratio]{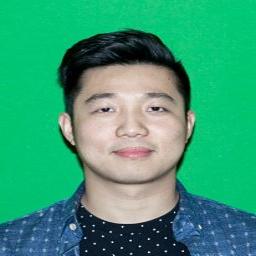}}]{Fan Zhu} is with Inception Institute of Artificial Intelligence (IIAI), Abu Dhabi, UAE.\
\end{IEEEbiography}

\vspace{-15mm}
\begin{IEEEbiography}[{\includegraphics[width=1in,height=1in,clip,keepaspectratio]{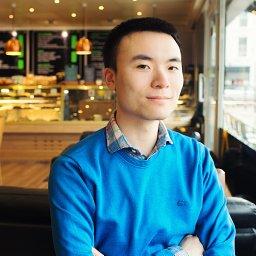}}]{Mengyang Yu} is a Research Scientist in the Inception Institute of Artificial Intelligence (IIAI), Abu Dhabi, UAE.\
\end{IEEEbiography}

\vspace{-15mm}
\begin{IEEEbiography}[{\includegraphics[width=1in,height=1in,clip,keepaspectratio]{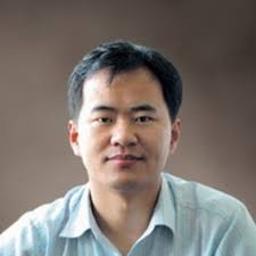}}]{Haoqian Wang} (M’13) received the B.S. and M.E. degrees from Heilongjiang University, Harbin, China, in 1999 and 2002, respectively, and the Ph.D. degree from the Harbin Institute of Technology, Harbin, in 2005. He was a Post-Doctoral Fellow with Tsinghua University, Beijing, China, from 2005 to
2007. He has been a Faculty Member with the
Tsinghua Shenzhen International Graduate School, Shenzhen, China, since 2008, where he has also been a Professor since 2018, and the director of Shenzhen Institute of Future Media Technology.
\end{IEEEbiography}

\vspace{-15mm}
\begin{IEEEbiography}[{\includegraphics[width=1in,height=1in,clip,keepaspectratio]{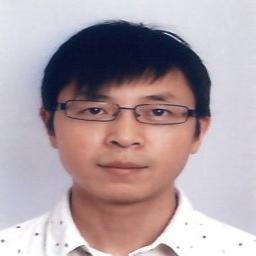}}]{Ling Shao} is the Executive Vice President and Provost of the Mohamed bin Zayed University of Artificial Intelligence. He is also the CEO and Chief Scientist of the Inception Institute of Artificial Intelligence (IIAI), Abu Dhabi, United Arab Emirates. His research interests include computer vision, machine learning, and medical imaging. He is a fellow of IAPR, IET, and BCS. 
\end{IEEEbiography}

\end{document}